\documentclass[10pt,twocolumn,letterpaper]{article}

\usepackage[pagenumbers]{cvpr} 

\usepackage{amssymb}
\usepackage{pifont}
\usepackage[dvipsnames]{xcolor}
\newcommand{\cmark}{\textcolor{ForestGreen}{\ding{51}}}%
\newcommand{\xmark}{\textcolor{Mahogany}{\ding{55}}}%

\usepackage{multirow}
\usepackage{fontawesome5}
\usepackage{colortbl}

\definecolor{cvprblue}{rgb}{0.21,0.49,0.74}
\usepackage[pagebackref,breaklinks,colorlinks,allcolors=cvprblue]{hyperref}
\usepackage[para]{footmisc}

\def\paperID{5762} 
\def\confName{CVPR}
\def\confYear{2026}

\title{RnG: A Unified Transformer for Complete  \\ 3D Modeling from Partial Observations}

\author{Mochu Xiang \textsuperscript{1,2\footnotemark[3]} \quad 
Zhelun Shen\textsuperscript{2\footnotemark[1] 
 \footnotemark[2]} \quad 
Xuesong Li\textsuperscript{3,4} \quad  
Jiahui Ren\textsuperscript{1} \quad 
Jing Zhang\textsuperscript{3} \\
Chen Zhao\textsuperscript{2} \quad 
Shanshan Liu\textsuperscript{2} \quad 
Haocheng Feng\textsuperscript{2} \quad 
Jingdong Wang\textsuperscript{2} \quad 
Yuchao Dai\textsuperscript{1\footnotemark[2]}\\
\small \textsuperscript{1}Northwestern Polytechnical University, China \quad
\textsuperscript{2}Baidu Inc., China \quad
\textsuperscript{3}Australian National University, Australia \quad
\textsuperscript{4}CSIRO, Australia 
\\
{\tt\small shenzhelun@pku.edu.cn \quad daiyuchao@nwpu.edu.cn}
}
\begin{document}
\twocolumn[{%
\renewcommand\twocolumn[1][]{#1}%
\begingroup
\hypersetup{linkcolor=black}
\maketitle
\endgroup
\centering
\vspace{-28pt}
\includegraphics[width=0.83\linewidth]{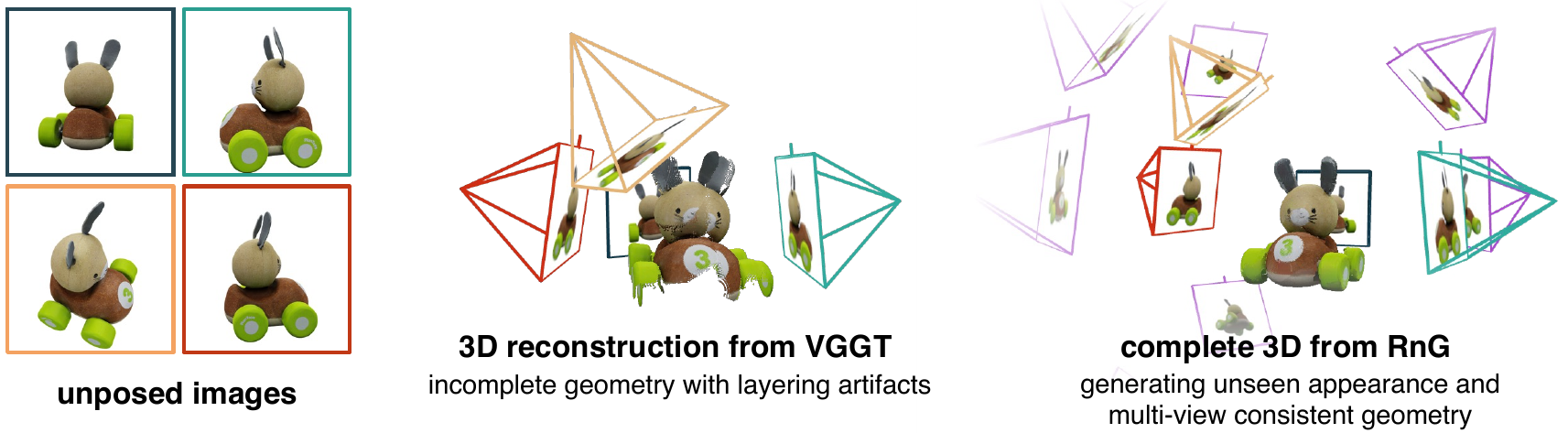}
\vspace{-6pt}
\captionof{figure}{\textbf{What can RnG do?} 
Given a few unposed images of an object, 3D reconstruction foundation models like VGGT can recover the structure of \emph{observed regions}, but leaves the unseen part un-modeled. RnG can estimate its complete 3D geometry within a second on an A800 GPU, using a single feed-forward transformer. RnG implicitly reconstructs 3D and render onto new viewpoints with appearance and geometry. By accumulating these rendered point maps , RnG can generate a complete 3D object, working like a virtual 3D scanner.
}
\label{fig:teaser}
\vspace{10pt}
}]

\begin{abstract}
Human perceive the 3D world through 2D observations from limited viewpoints. While recent feed-forward generalizable 3D reconstruction models excel at recovering 3D structures from sparse images, their representations are often confined to observed regions, leaving unseen geometry un-modeled. This raises a key, fundamental challenge: \emph{Can we infer a complete 3D structure from partial 2D observations?}
We present RnG (Reconstruction and Generation), a novel feed-forward Transformer that unifies these two tasks by predicting an implicit, complete 3D representation. At the core of RnG, we propose a reconstruction-guided causal attention mechanism that separates reconstruction and generation at the attention level, and treats the KV-cache as an implicit 3D representation. Then, arbitrary poses can efficiently query this cache to render high-fidelity, novel-view RGBD outputs. As a result, RnG not only accurately reconstructs visible geometry but also generates plausible, coherent unseen geometry and appearance.
Our method achieves state-of-the-art performance in both generalizable 3D reconstruction and novel view generation, while operating efficiently enough for real-time interactive applications. 
Project page: \url{https://npucvr.github.io/RnG}
\vspace{-16pt}
\end{abstract}

\makeatletter
\renewcommand{\thefootnote}{\fnsymbol{footnote}}
\makeatother
\footnotetext[1]{Project leader.\hspace{10pt}}
\footnotetext[2]{Corresponding authors.\hspace{80pt}}
\footnotetext[3]{Work done during an internship at Baidu.}
\section{Introduction}
\label{sec:intro}

3D perception from sparse visual data is a cornerstone of modern computer vision, supporting advancements in robotics~\cite{kitti}, augmented reality~\cite{cvd}, and large-scale digital content creation~\cite{trellis}. Recent progress in generalizable 3D reconstruction~\cite{dust3r,vggt} has demonstrated the feasibility of recovering scene geometry and appearance from just a few unposed input images. However, a fundamental limitation persists across most reconstruction pipelines: these models are explicitly trained to recover geometry only for regions visible in the input camera views. Consequently, their geometric outputs are incomplete, hindering downstream applications that demand a complete object representation, such as physics simulation~\cite{dso} or content generation~\cite{syncity}.

Novel view synthesis (NVS) extends these capabilities by generating images from unseen viewpoints based on observed data. While certain NVS pipelines exhibit generalization to unobserved regions, they typically lack consistent 3D structure. As \autoref{tab:intro_comparison} summarizes, these generalized NVS methods can render photorealistic novel views but are often limited by their dependence on posed inputs, lack of geometric reconstruction, or restricted camera control. Recently, Matrix3D~\cite{matrix3d} explores a unified reconstruction and novel view generation approach. However, its diffusion-based design incurs high computational costs, making it unsuitable for real-time applications.

Despite these challenges, we hypothesize that feed-forward 3D reconstruction foundation models~\cite{dust3r,vggt}, while primarily trained to recover visible geometry, may possess the potential to infer a more complete 3D representation within their latent space. If this potential can be effectively harnessed and then rendered through viewpoint-conditioned neural decoding, it could reveal the model's internal, comprehensive 3D understanding.

We introduce \textbf{RnG}, a feed-forward transformer designed to activate and make this latent perception explicit. By treating viewpoint-conditioned neural rendering as a query to the model's latent features, RnG bridges reconstruction and generation within a single, unified framework. Specifically, to handle unposed input while maintaining explicit camera control, we build on 3D reconstruction foundation models~\cite{vggt} and inherit their meaningful latent 3D representations. Our single model jointly reconstructs latent 3D and generates novel view appearance and geometry. This is achieved via a reconstruction-guided causal attention mechanism, which decouples these two processes at the attention level while using the same set of parameters for both tasks. It represents a new direction that transfers knowledge from 3D reconstruction to image generation. Although prior work has focused on adapting generative priors for reconstruction~\cite{marigold, matrix3d}, we demonstrate that the reverse direction of transferring reconstruction priors to generation is viable and effective.

\begin{table}[t]
\centering
\small
\setlength\tabcolsep{2pt}
\caption{Comparison between representative 3D reconstruction and novel view synthesis methods.}
\resizebox*{1.0\linewidth}{!}{
\begin{tabular}{l l|cc|ccc|cc}
\toprule
\multicolumn{2}{c|}{Task} & \multicolumn{2}{c|}{3D reconstruction} & \multicolumn{3}{c|}{Novel view synthesis} & \multicolumn{2}{c}{Unified model} \\
\midrule
\multicolumn{2}{c|}{\multirow{2}{*}{Method}} & VGGT & DUSt3R & LVSM & LGM & RayZer & Matrix3D & RnG \\
\multicolumn{2}{c|}{}                         & \cite{vggt} & \cite{d2ust3r} & \cite{lvsm} & \cite{lgm} & \cite{rayzer} & \cite{matrix3d} & Ours \\
\midrule
\multirow{2}{*}{\faCamera} 
& unposed infer & \cmark & \cmark & \xmark & \xmark & \cmark & \cmark & \cmark \\
& camera control     & N/A & N/A & \cmark & \cmark & \xmark & \cmark & \cmark \\
\midrule
\multirow{2}{*}{\faCube}  
& generate unseen & \xmark & \xmark & \cmark & \xmark & \cmark & \cmark & \cmark \\
& explicit 3D        & \cmark & \cmark & \xmark & \cmark & \xmark & \cmark & \cmark \\
\midrule
\faClock & real-time infer                 & \cmark & \cmark & \cmark & \cmark & \cmark & \xmark & \cmark \\
\bottomrule
\end{tabular}
} 
\label{tab:intro_comparison}
\end{table}

To generalize to unobserved viewpoints, we adopt an implicit formulation: novel views are used as queries to the latent 3D space, and the network's generation stage \textit{renders} the implicit 3D onto those viewpoints. Additionally, we render a point map at each queried viewpoint, converting the implicit latent 3D to explicit geometric representations. Since RnG is a deterministic feed-forward transformer, it provides real-time inference capability. The causal attention design enables a KV-Cache mechanism, allowing the model to \textit{\textbf{R}econstruct} from input images once (in $\sim$0.2s) and subsequently \textit{\textbf{G}enerate} novel-view appearances and geometry with high efficiency (in $<$ 0.1s). RnG runs over $100\times$ faster than its diffusion-based counterparts~\cite{matrix3d}.

Experimentally, RnG achieves state-of-the-art performance across multiple 3D reconstruction and novel view generation benchmarks, including camera pose estimation, input and novel view depth prediction and novel view synthesis. Remarkably, it surpasses specialized models designed for individual tasks by a significant margin.
We summarize the key highlights of RnG as follows:

\begin{itemize}
    \item \textbf{Unified architecture via Causal Attention.} We introduce a reconstruction-guided causal attention mechanism. This decouples the two tasks at the attention level, allowing a single feed-forward transformer to coherently perform implicit 3D reconstruction, novel view synthesis, and explicit geometry generation.

    \item \textbf{KV-Cache as an Implicit 3D Representation.} Enabled by our causal attention design , the network’s KV-Cache serves as a flexible implicit 3D representation. This facilitates highly efficient novel view generation (rendering) and provides a transparent link between internal memory and reconstructed geometry. 

    \item \textbf{Reconstruction-driven generation.} RnG demonstrates that leveraging reconstruction priors, rather than generative ones, enables superior novel view synthesis performance while significantly reducing computational cost.

\end{itemize}
\section{Related Work}
\label{sec:related_work}

\subsection{Neural Rendering}

Neural rendering represents a scene by mimicking how cameras capture the real world~\cite{neural_rendering_survey}. Some differentiable rasterization~\cite{nvdiffrast} or ray-tracing\cite{DrJit} methods directly formulate the 3D-to-2D imaging process from 3D meshes. 
To recover the scene from image collections, radiance fields can provide photorealistic images. These representations are encoded in neural network weights~\cite{nerf}, voxel grids~\cite{plenoxels} or hash-grids~\cite{instant_ngp}.
3D Gaussian Splatting~\cite{gs} explicitly model the scene using 3D Gaussian primitives that are similar to point clouds.
To get a novel view image, all these methods require a physically explainable rendering process. More recently, there are attempts that directly use a neural network to generate novel view images, whether from multi-view images~\cite{srt, lvsm} or from 3D meshes~\cite{renderformer}.

\subsection{Novel View Synthesis}

\begin{figure*}
\centering
\small
\setlength{\tabcolsep}{4pt}
\resizebox{\linewidth}{!}{
\begin{tabular}{c !{\color{gray}\vrule} c}
    {\includegraphics[height=0.39\linewidth]{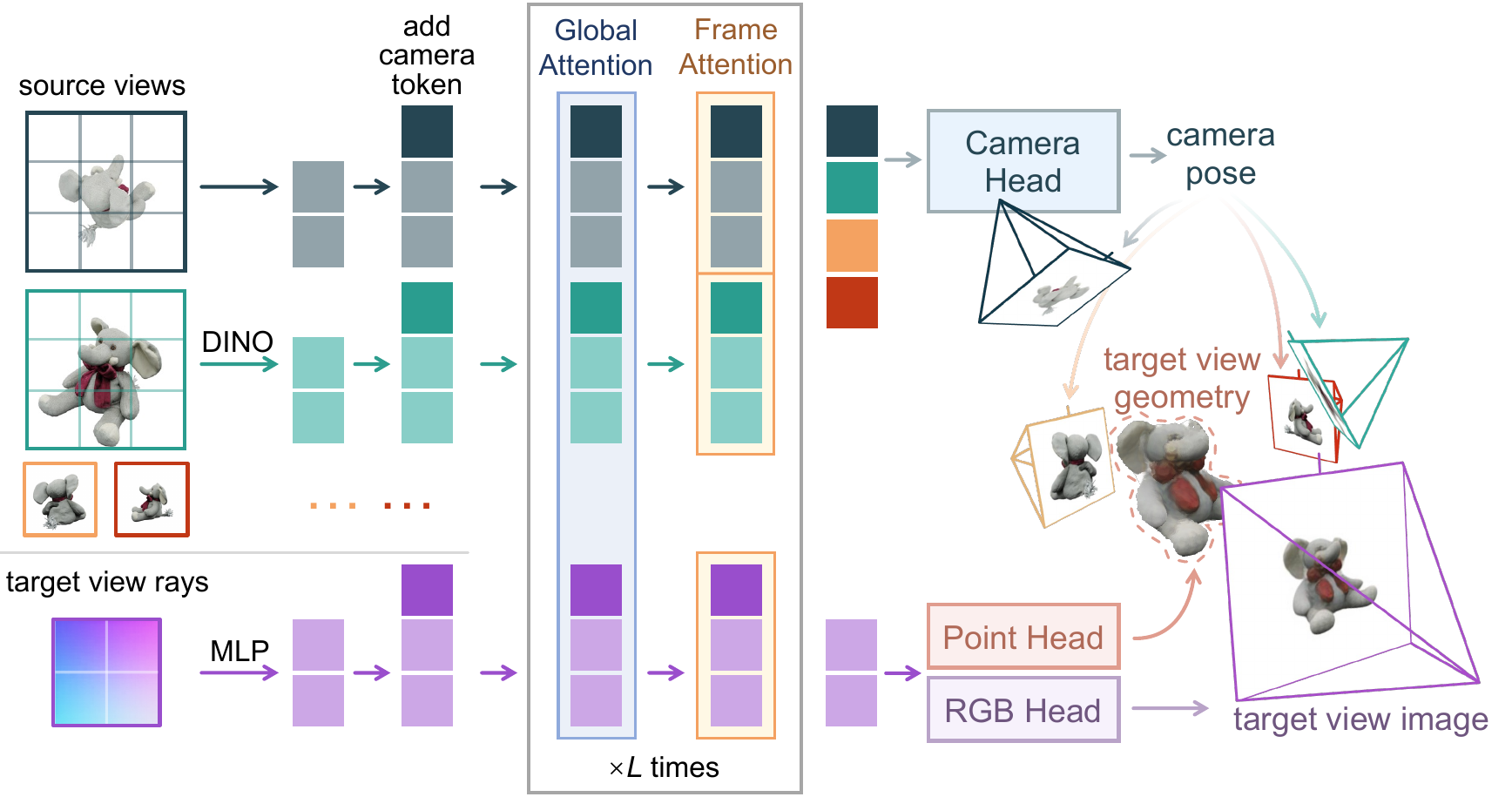}} &
    {\includegraphics[height=0.39\linewidth]{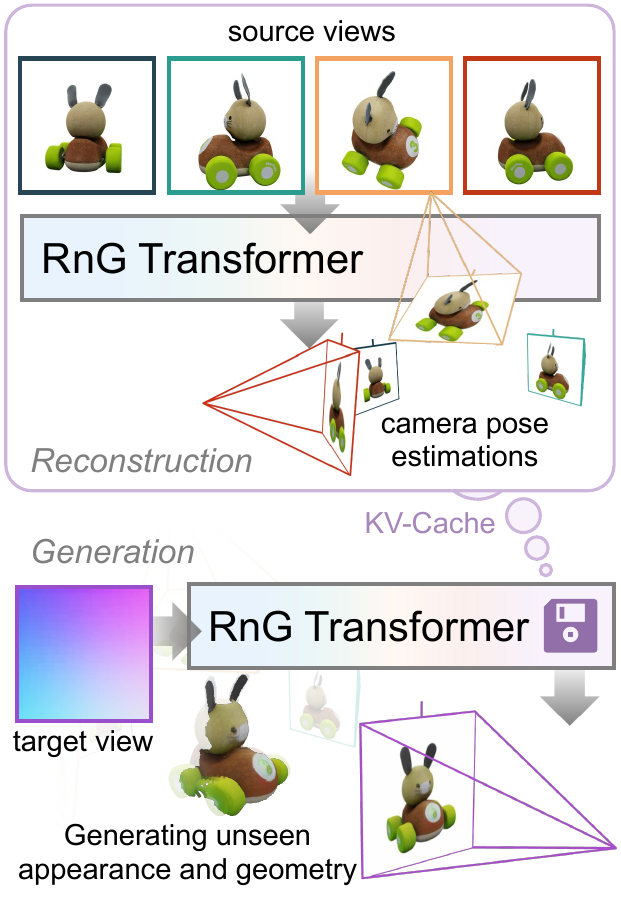}} \\ 
    (a) The model architecture of RnG &
    (b) The causal inference of RnG \\
\end{tabular}
} 
\caption{\textbf{The Network Architecture of RnG.} (a) Source view images are first tokenized using the DINO vision transformer; the Plücker ray map representing the target view point goes through a linear layer. After adding camera tokens for each view, all tokens will then alternately attend to global- and frame-level attention blocks. Finally, camera tokens from input views are used to estimate camera poses, while a point head and an RGB head process ray tokens from the target view, providing geometry and appearance estimations. (b) In inference, the model can cache K/V token from source views, synthesizing novel view geometry and geometry at a higher speed.
}
\label{fig:architecture}
\end{figure*}

Novel View Synthesis (NVS) aims to generate images at novel viewpoints based on given observations. This task is typically approached in two distinct settings. When dense observations with sufficient coverage are available, methods leveraging implicit~\cite{nerf, instant_ngp} or explicit~\cite{gs, nvdiffrast} representations can overfit to a specific scene, enabling high-fidelity novel views through neural rendering or rasterization.

In the more challenging sparse-observation setting, generalizable methods are required. Diffusion-based approaches have shown promise in generating diverse novel views, but often struggle to maintain strict 3D consistency~\cite{zero1to3}. Subsequent works have sought to improve multi-view consistency through implicit~\cite{eschernet} or explicit~\cite{consistnet} 3D modeling. This line of work also includes video-diffusion models~\cite{svc, recammaster} for dynamic scenarios and multi-modal methods~\cite{matrix3d, aether} that generate explicit 3D geometry (e.g., posed-RGBD) alongside appearance.

As an alternative to costly diffusion models, feed-forward methods offer significantly faster inference. One popular strategy involves adapting 3D Gaussian Splatting for sparse inputs~\cite{pixelsplat, mvsplat, depthsplat, hisplat, mvsplat360, nopesplat}. To better synthesize occluded regions, other implicit methods~\cite{srt, lvsm} use a neural network to directly estimate the novel view image. These approaches often incorporate specialized network designs, such as multi-view attention mechanisms~\cite{gta, rrope} or epipolar geometry encodings~\cite{pixelnerf, gpnr}. While many of these methods require known camera poses, a distinct sub-class of unposed methods~\cite{rust, rayzer} operates without known pose, though their estimated latent poses are less interpretable. 

\subsection{3D Vision Foundation Models}

Dust3R~\cite{dust3r} and VGGT\cite{vggt} demonstrate effective data-driven solutions to 3D computer vision tasks. Inspired by their practice, recent attempts benefit from their extensive pretraining data and demonstrate superior performance in wide range of tasks.
Attaching a Gaussian parameter prediction head can transform the point map prediction into photo-realistic scene reconstructions. Several attempts explored object-level~\cite{freesplatter} and scene-level~\cite{splatt3r, nopesplat} adaptations.
Extending model inputs to more views~\cite{fast3r, mvdust3r} or adapting models to accept streaming inputs~\cite{cut3r, stream3r2025} expand the model's application to larger scales.
Extensive training data also benefits the domain adaptation process, bringing greater performance in other domains~\cite{aerialmegadepth, endo3r, panosplatt3r} or in model prompting~\cite{pow3r}.
Although the training data mainly covers static scenes, they can be easily adapted to perform dynamic reconstructions~\cite{stereo4d, easi3r, monst3r, align3r, pi3}.

Generative priors~\cite{stablediffusion} have proven highly effective in transfer learning. Remarkable progress are made in repurposing diffusion-based image generation backbones in dense prediction tasks~\cite{marigold} and multi-modal learning~\cite{matrix3d, aether, zhang2025world}. Although employing generative priors can effectively aid the perception task, the diffusion-based denoising paradigm greatly increases inference time, imposing a substantial burden on real-time interactive applications.

\section{Method}
Given a few unposed source images $\{\mathbf I_s\}$, RnG reconstructs the geometry of the source views and synthesizes novel-view geometry and appearance.
Specifically, RnG first performs implicit 3D reconstructions, stored in network's KV-Cache. Meanwhile, the camera poses $\{\hat{\mathbf g}_s\}$ for source images are estimated for later use. Then, RnG reads from this KV-Cache to efficiently render at an arbitrary viewpoint $\mathbf g_t$, providing novel appearance $\hat{\mathbf I}_t$ and a point map $\hat{\mathbf{p}}_t$. By accumulating structures queried from multiple viewpoints or directly mapping textures according to the estimated source views pose $\hat {\mathbf g}_s$, a complete 3D structure can be generated. The entire process only takes a second.

To achieve this, we begin by outlining the overall architecture of RnG, as described in Section~\ref{sec:model_architecture}. 
RnG performs source-view reconstruction and novel-view generation using a single network; we design a reconstruction-guided causal attention, where the two processes are decomposed at the attention level (see Section~\ref{sec:attention_decompose}). 
Furthermore, this design allows the network’s KV-Cache to serve as an implicit 3D representation (Section~\ref{sec:KV-Cache}). 

\subsection{A Reconstruction and Generation Unified Transformer}
\label{sec:model_architecture}

Our goal is to recover 3D structure and appearance at arbitrary viewpoints from a limited set of unposed images. This requires us to bridge the 3D reconstruction and novel view generation with a shared implicit representation.
To this end, we propose a Reconstruction-and-Generation unified transformer (RnG). The overall architecture is shown in Fig.~\ref{fig:architecture}, we inherit the model architecture and the weights of VGGT~\cite{vggt}. As illustrated, an image-level feature extractor~\cite{dinov2} first converts the source-view images $\{\mathbf I_s\}$ into tokens. We then encode the target view as a Plücker ray map and project it through a linear layer to obtain target-view tokens. Tokens from both the source and target views are concatenated with their respective camera and register tokens, and the resulting sequence is processed by interleaved global and frame attention layers.

After $L=24$ layers of interleaved global and frame attention, the source-view tokens are used by a camera head to predict camera poses $\{\hat{\mathbf g}_s\}$. For the target-view tokens, we attach two DPT heads~\cite{dpt} to separately generate novel view appearance and geometry. Specifically, tokens derived from the target view’s Plücker ray map (encoded from $\mathbf g_t$) are fed to an RGB head $\mathcal D_\text{RGB}$ to produce a novel view image $\hat{\mathbf I}_t$, while a point head $\mathcal D_\text{pmap}$ outputs a pixel-aligned point map $\hat{\mathbf p}_t$ representing the corresponding region's geometry. 

Following the practice of VGGT~\cite{vggt} and better preserve its learned knowledge, we assign the first source view with special camera and register tokens, and let the rest source views and the target view share the same ones. We train the model to always regress a fixed location for the first view:
\begin{align}
    \hat {\mathbf g}_{s=1} = \left[I_{3\times3} \left| [0, 0, -1]^\intercal \right. \right],
\end{align}
which implicitly assigns a world coordinate for reconstructions. 
The target viewpoint $\mathbf g_t$ and the estimated point map $\hat {\mathbf p}_t$ are also expressed in this world coordinate system.

\subsection{Reconstruction-Guided Causal Attention}
\label{sec:attention_decompose}

\begin{figure}
    \centering
    \includegraphics[width=\linewidth]{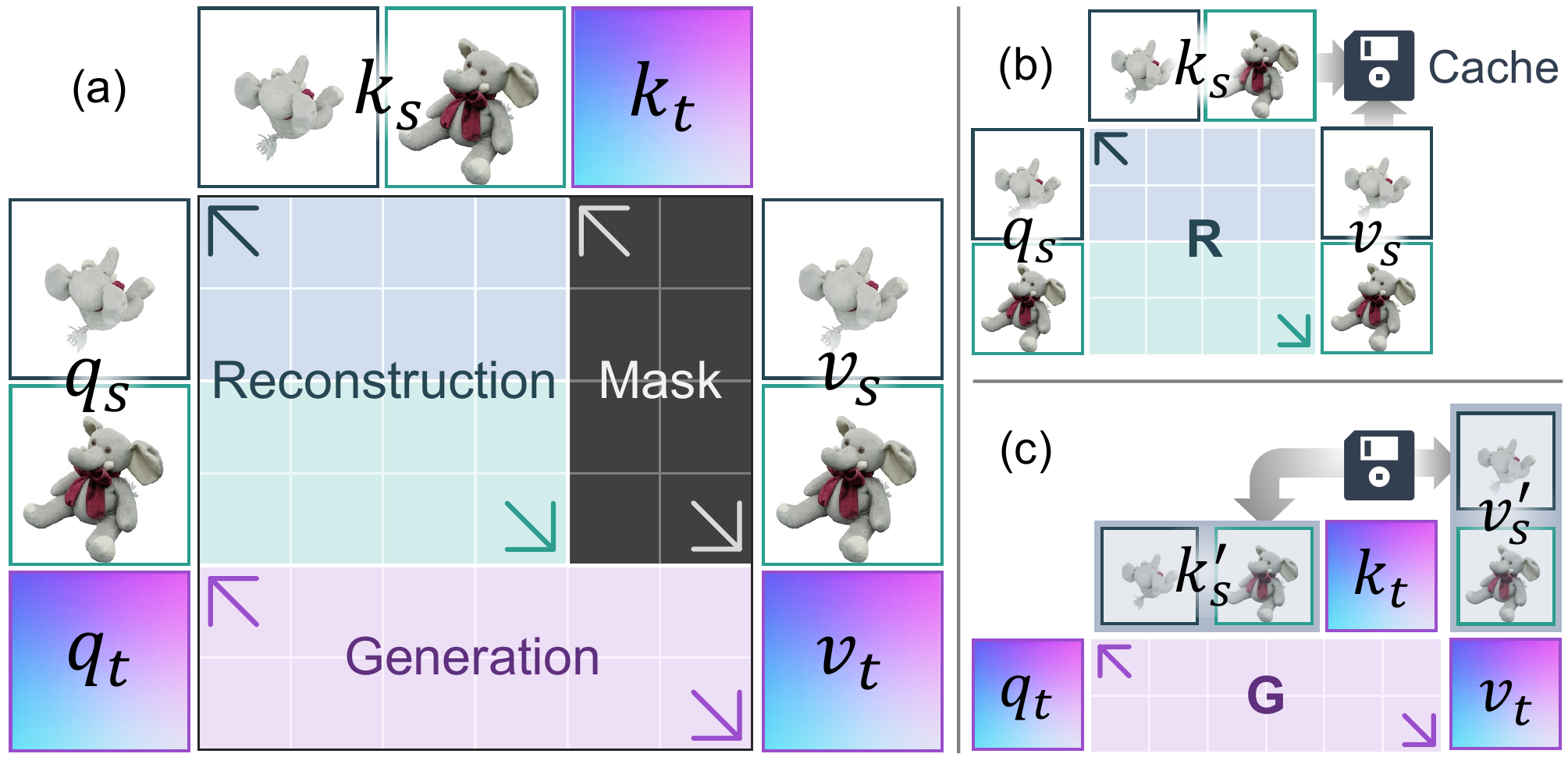}
    \caption{\textbf{The reconstruction-guided causal attention.} \textbf{(a)} During training, we decouple reconstruction and generation at the attention level inside global attention blocks. At inference time, the attention process is split into two steps: \textbf{(b)} source-view key value tokens are cached as an implicit 3D representation; \textbf{(c)} the KV-cache is queried by target view poses to generate novel views.
    }
    \label{fig:mask_attention}
\end{figure}

Intuitively, reconstruction should guide the generation process, but generation should not interfere with reconstruction. This will encourage consistent perceptions for source views and leverage the geometric understanding encoded in the latent features to enhance novel-view generation.

Inspired by the causal attention used in language~\cite{touvron2023llama, gpt3} and vision~\cite{yang2021causal, zhuo2025streaming} models, we proposed reconstruction-guided causal attention to achieve this goal. Specifically, we introduce a binary mask $M$ to control the information flow during attention computation:
\begin{equation}
M_{i,j} = \left\{
\begin{array}{c l}
0 & \text{if} \ i\in \{s\} \ \text{and} \ j \in \{t\}  \\
1 & \text{elsewhere} \\
\end{array}
    \right. ,
\end{equation}
where $\{s\}$ and $\{t\}$ are indexes of tokens from source views and target views. This results in the attention output ($d_k$ is the feature dimension of query and key tokens):
\begin{equation}
    \text{Out} = \text{softmax} \left( \frac {M \odot Q K^\intercal} {\sqrt{d_k}} \right) V .
\end{equation}

~\autoref{fig:mask_attention}~(a) illustrates the proposed mechanism in detail. It forbids query tokens from source views attending to keys from the target view. In other words, query tokens from source views can only attend to keys from source views, whereas target view tokens can see those from both source views and target views. By using masked attention, our model can provide consistent reconstruction results when given arbitrary target views.

It is worth noting that the network uses shared parameters to process tokens from both source views and target views, even though they respond to different aspects of the task: tokens from source views are responsible for perception and estimating camera poses (reconstruction), while tokens from the target view are responsible for synthesizing novel appearance and geometry (generation). This requires us to joint train the network to perform two tasks. The reconstruction-guided causal attention helps decompose the two tasks at the attention level, while keeping our solution parameter-efficient.

Crucially, this causal design is not just a mechanism for task separation; it directly enables a highly efficient, two-stage inference process and allows the network's KV-Cache to be re-interpreted as a complete, implicit 3D representation, as detailed in the following section.

\subsection{KV-Cache as an Implicit 3D Representation}
\label{sec:KV-Cache}
Another important property of reconstruction-guided causal attention is that it allows us to reinterpret the cached key/value tokens of each attention block as an implicit 3D representation: a latent memory space that encodes scene geometry and appearance, independent of the viewing direction. This allows us to decouple the inference process into two distinct stages, as illustrated in~\autoref{fig:mask_attention}(b) and (c):

\noindent\textbf{Stage 1: Reconstruction and Caching.} Because the attention processes for source-view tokens are independent of the existence of target-view tokens, the model can, at inference time, reconstruct scenes using only the source views, without requiring any target view. Specifically, we cache the key and value tokens from the source views in each global attention block, which then serve as reusable scene representations for subsequent novel-view generation.
\begin{equation}
K_s' = \text{Cache}(K_s),\quad V_s' = \text{Cache}(V_s) .
\end{equation}

\noindent\textbf{Stage 2: Generation and Querying.} Then, to synthesize novel views, we can simply forward target view Pl\"ucker rays without computing any global attention and frame attention for source views, but directly read from the cache:
\begin{equation}
\text{Out}_t = \text{softmax}\left(\frac{Q_t \cdot [K_s'; K_t]^\intercal} {\sqrt{d_k}} \right) [V_s'; V_t],
\end{equation}
where $[;]$ denotes concatenating along the token dimension. This two-stage approach significantly reduces the computation and speeds up the rendering process.

After $L$ layers of interleaved global and frame attention, the target-view tokens $\text{Out}_t$ are decoded by DPT heads~\cite{dpt} $\mathcal D_\text{RGB}$ and $\mathcal D_\text{pmap}$ to predict an RGB image and a point map:
\begin{equation}
    \hat{\mathbf I} = \mathcal D_\text{RGB} (\text{Out}_t) , \quad
    \hat{\mathbf P} = \mathcal D_\text{pmap} (\text{Out}_t) .
\end{equation}
Finally, we can recover the full 3D structure by accumulating structures from multiple viewpoint queries.

\subsection{Training strategy}

We train RnG using a multi-task loss:
\begin{equation}
    \mathcal L = \mathcal L_{RGB} + \lambda_{pmap} \mathcal L_{pmap} + \lambda_c\mathcal L_{cam}.
\end{equation}

To supervise the novel view image, we use a combination of MSE loss and perceptual loss~\cite{johnson2016perceptual} between the ground truth $\mathbf I_t$ and the rendered view $\hat {\mathbf I}_t$:
\begin{equation}
    \mathcal L_{RGB} = \left| \mathbf I_t - \hat {\mathbf I}_t \right|_2 + \lambda_\text{p} \cdot \text{Perceptual}\left(\mathbf I_t, \hat {\mathbf I}_t\right).
\end{equation}

The point head decodes a four-channel map: a point map $\hat{\mathbf P}_t$ representing the $xyz$-locations for visible regions and a single-channel uncertainty map $\Sigma_t$, supervised by an aleatoric-uncertainty loss~\cite{aleatoric_uncertainty}, following~\cite{dust3r, vggt}:
\begin{equation}
\begin{split}
    \mathcal L_{pmap} =& \left\| \Sigma_t \odot (\mathbf P_t - \hat {\mathbf P}_t) \right\| \\
    +& \left\| \Sigma_t \odot (\nabla \mathbf P_t - \nabla \hat {\mathbf P}_t) \right\| - \alpha \cdot \log \Sigma_t.
\end{split}
\end{equation}

The estimated camera poses for input frames $\hat{\mathbf g}_s$ are supervised using the Huber loss:
\begin{equation}
    \mathcal L_{cam} = \sum_s \left|\mathbf g_s - \hat {\mathbf g}_s \right|_{\epsilon}.
\end{equation}
Here, the ground truth camera poses in one batch are normalized so that ${\mathbf g}_{s=1} = \left[I_{3\times3} \left| [0, 0, -1]^\intercal \right. \right]$.

\section{Experiments}
\label{sec:experiments}

\begin{table*}[t]
\centering
\caption{\textbf{Quantitative comparison.} We evaluate the reconstruction and generation ability of all models on the GSO dataset. `—' means that the model is not capable of delivering that result.}
\setlength{\tabcolsep}{4pt}
\resizebox{\linewidth}{!}{
\begin{tabular}{ c  c | ccc | cc | cc | ccc | c} \toprule
 & & \multicolumn{5}{c|}{\textit{Reconstruction}} & \multicolumn{5}{c|}{\textit{Generation}} & \textit{Complete} \\
 
 & & \multicolumn{3}{c|}{Pose} &
   \multicolumn{2}{c|}{Source View Depth} & 
   \multicolumn{2}{c|}{Novel View Depth} & 
   \multicolumn{3}{c|}{Novel View Synthesis} &
   \textit{3D}
   \\

 & & RA@5$\uparrow$ & RT@5$\uparrow$ & AUC@30$\uparrow$ &
     Rel$\downarrow$ & a1$\uparrow$ & Rel$\downarrow$ & a1$\uparrow$ &
     PSNR$\uparrow$ & SSIM$\uparrow$ & LPIPS$\downarrow$ & CD$\downarrow$ \\ 
    
\midrule

 \multirow{4}{*}{\textit{posed}} &
 LVSM\cite{lvsm}  & \multicolumn{3}{c|}{\multirow{4}{*}{{\textit{camera poses are known}}}} &
 --- & --- & --- & --- & 
\textbf{ 27.522} & \textbf{0.902} & \textbf{0.0895} 
& --- \\
 
 & LGM\cite{lgm}  & & & &  
 13.704 & 85.646 & 16.804 & 82.205 & 
 16.677 & 0.751 & 0.2248 
 & 0.1130 \\
 
 & Matrix3D\cite{matrix3d}  & & & &  
 ~~8.782 & 95.383 & ~~8.897 & 93.858 & 
 19.941 & 0.803 & 0.1684 
 & 0.0580 \\ 
  
 & pixelSplat\cite{pixelsplat} & & & & 
 33.594 & 33.241 & 39.819 & 20.723 & 
 12.904 & 0.718 & 0.4514 & 0.4898\\
 
 \midrule
 
 \multirow{5}{*}{\textit{unposed}} 
 & MUSt3R\cite{must3r} & 21.683 & 21.683 & 33.490 &
 10.941 & 92.663 & — & — & — & — & — & 0.7757 \\
 
& VGGT\cite{vggt}  & 74.239 & 65.680 & 77.234 &
             ~~5.960 & 97.723 & — & — & 
             — & — & — 
              & 0.0260 \\
              
& VGGT\cite{vggt} + LVSM\cite{lvsm}  & 74.239 & 65.680 & 77.234 &
             ~~5.960 & 97.723 & — & — & 
             19.394 & 0.796 & 0.1980 
              & 0.0260 \\
             
 & Matrix3D\cite{matrix3d}  & 43.770 & 65.922 & 66.389 & 
               ~~9.430 & 92.261 & ~~9.964 & 90.282 & 
               18.736 & 0.786 & 0.1930 
                & 0.0670\\
               
 & RnG (Ours) & \textbf{85.146} & \textbf{86.019} & \textbf{86.942} & 
          ~~\textbf{0.584} & \textbf{99.929}  & ~~\textbf{0.717} & \textbf{99.850} & 
         26.276 & 0.891 & 0.0975 
          & \textbf{0.0067} \\
\bottomrule

\end{tabular}
} 
\label{tab:main_comparison}
\end{table*}

\begin{figure*}
\centering
\small 
\setlength{\tabcolsep}{4pt}
\resizebox{\linewidth}{!}{
\begin{tabular}{c  c c c c c c} \toprule
& \multicolumn{2}{c}{\textit{posed}} & \multicolumn{3}{c}{\textit{unposed}} \\ \cmidrule(lr){2-3} \cmidrule(lr){4-6}
\multicolumn{1}{c}{Source Views} & LVSM\cite{lvsm} & LGM\cite{lgm} & Matrix3D\cite{matrix3d} & V\cite{vggt}+L\cite{lvsm} & RnG (Ours) & GT \\ \midrule

{\includegraphics[width=0.15\linewidth]{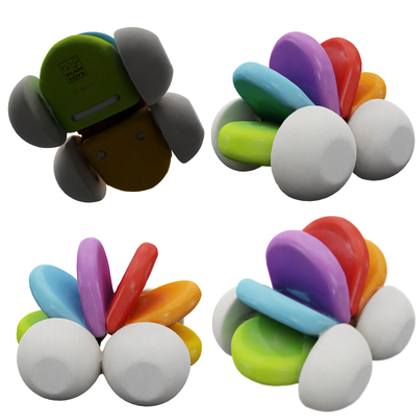}} & 

{\includegraphics[width=0.15\linewidth]{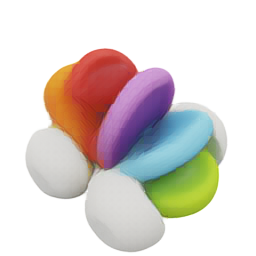}} & 
{\includegraphics[width=0.15\linewidth]{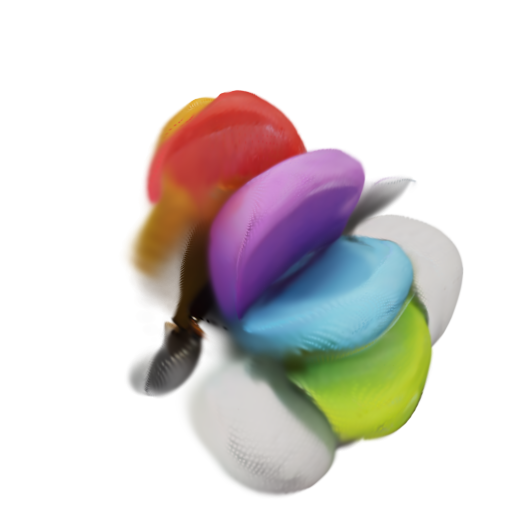}} & 
{\includegraphics[width=0.15\linewidth]{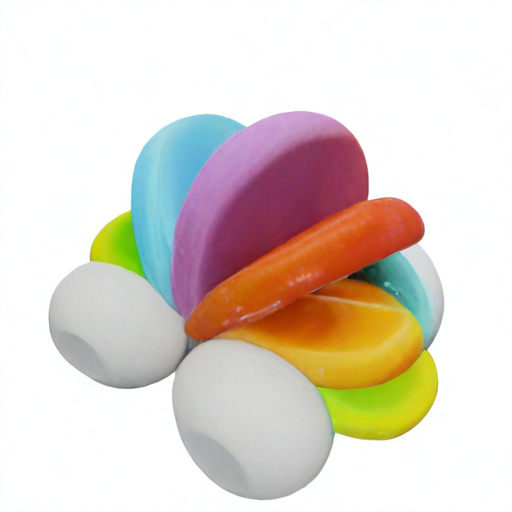}} & 
{\includegraphics[width=0.15\linewidth]{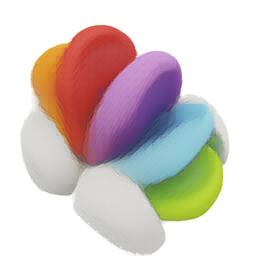}} & 
{\includegraphics[width=0.15\linewidth]{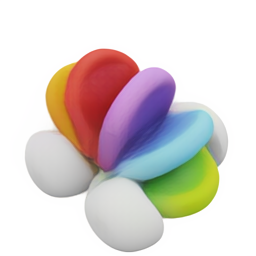}} & 

{\includegraphics[width=0.15\linewidth]{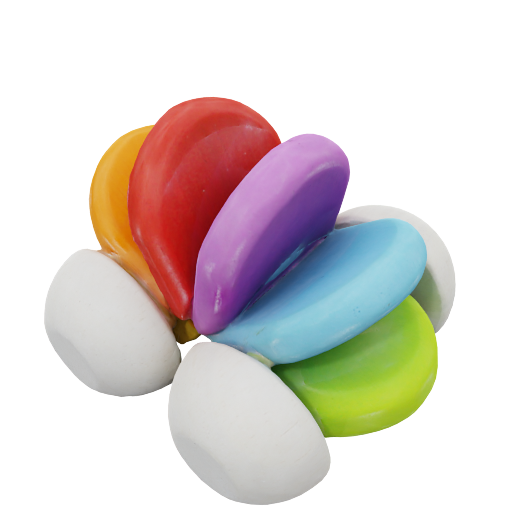}} \\

{\includegraphics[width=0.15\linewidth]{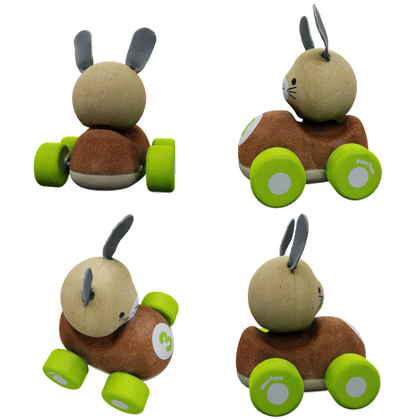}} & 

{\includegraphics[width=0.15\linewidth]{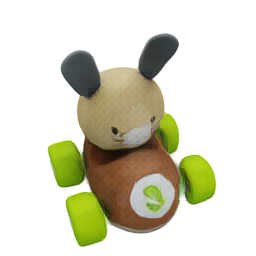}} & 
{\includegraphics[width=0.15\linewidth]{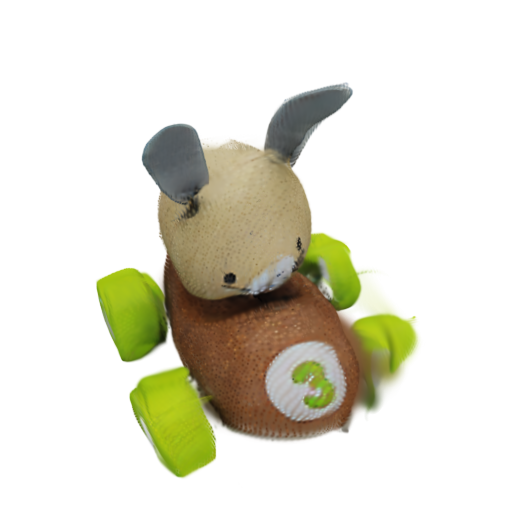}} & 
{\includegraphics[width=0.15\linewidth]{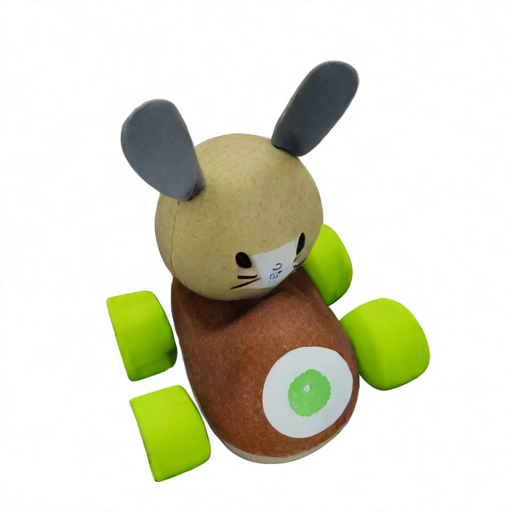}} & 
{\includegraphics[width=0.15\linewidth]{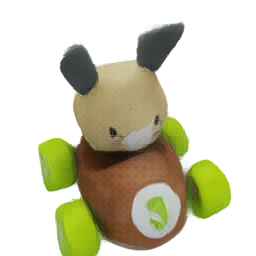}} & 
{\includegraphics[width=0.15\linewidth]{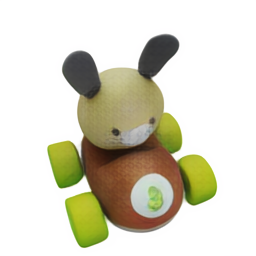}} & 

{\includegraphics[width=0.15\linewidth]{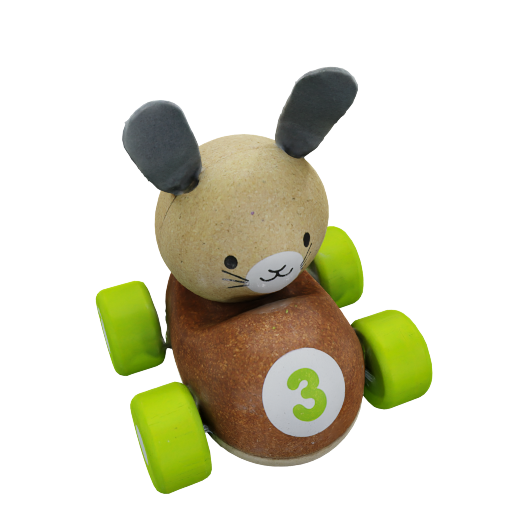}} \\

{\includegraphics[width=0.15\linewidth]{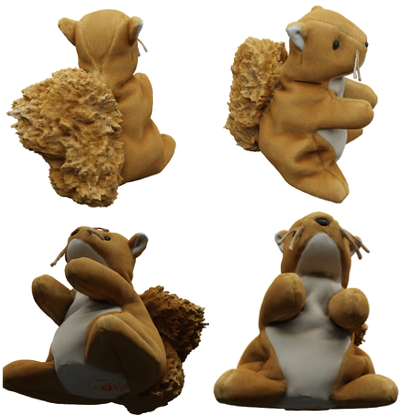}} & 

{\includegraphics[width=0.15\linewidth]{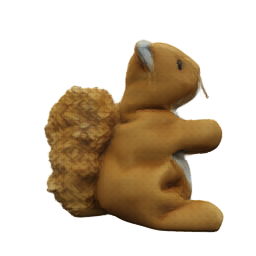}} & 
{\includegraphics[width=0.15\linewidth]{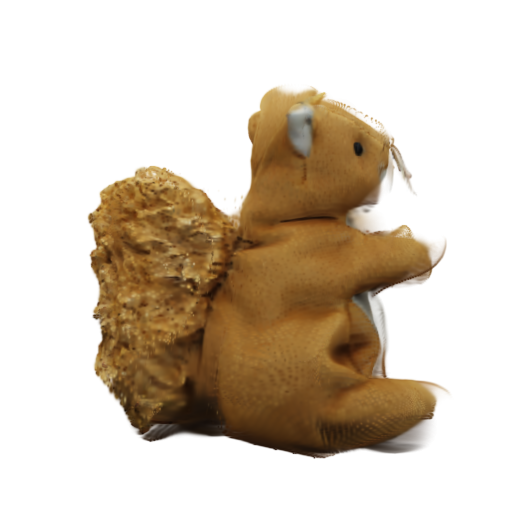}} & 
{\includegraphics[width=0.15\linewidth]{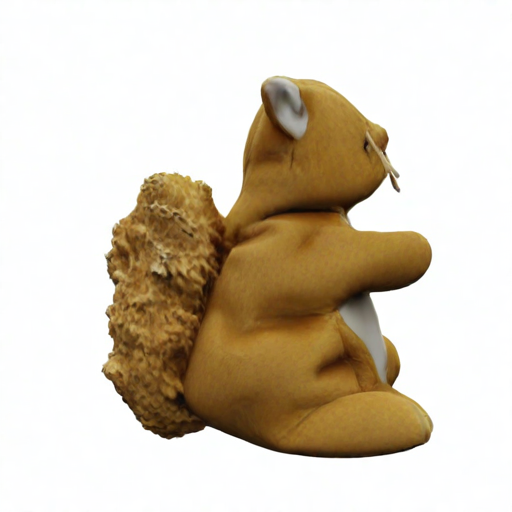}} & 
{\includegraphics[width=0.15\linewidth]{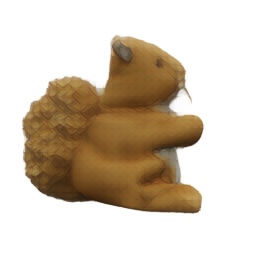}} & 
{\includegraphics[width=0.15\linewidth]{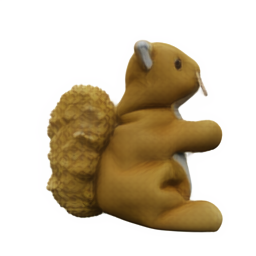}} & 

{\includegraphics[width=0.15\linewidth]{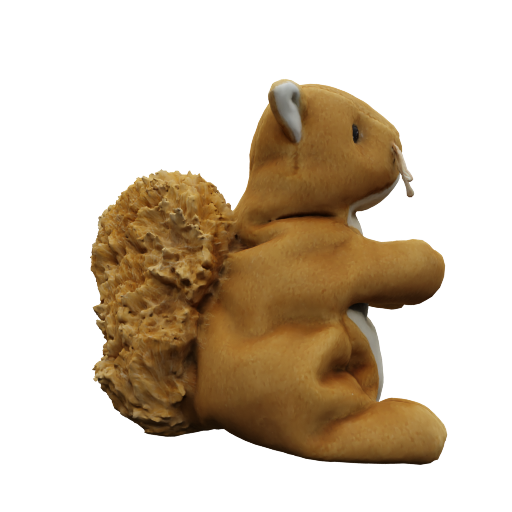}} \\ 

\bottomrule
     
\end{tabular}
} 
\caption{\textbf{Visual comparison of novel view synthesis.} Though RnG does not require accurate pose as input, it provides comparable visual quality with state-of-the-art pose-dependent methods like LVSM. Our model can hallucinate unseen regions with high 3D consistency.}
\label{fig:main_visual}
\end{figure*}

\subsection{Experimental Details}

\noindent\textbf{Dataset.}
We use 3D object datasets to get multi-view RGBD renderings. For model training, we use the Objaverse\cite{objaverse} dataset. We use the filtered list from LVIS subset and LGM\cite{lgm}, resulting in a total of 113.5K objects. For evaluation, we use the Google Scanned Objects (GSO)~\cite{gso_dataset} dataset, which contains 1030 objects. We normalize the 3D object and put it at the world origin. We then randomly place cameras around the object, where all cameras look at the world origin, but at varying distances.

\noindent\textbf{Comparing Methods.}
We compare our method with state-of-the-art approaches in novel view synthesis and 3D reconstruction, including methods that require accurate camera poses as input~\cite{lvsm, lgm, matrix3d} and unposed methods~\cite{matrix3d}. 
Since the object-level model of LVSM is not publicly available, we train LVSM using the official implementation on our own training data for a fair comparison. 

\noindent\textbf{Evaluation Metrics.}
We evaluate models on reconstruction metrics and novel view generation metrics. Reconstruction metrics are assessed on input views, where camera pose and depth are evaluated. Generation metrics are evaluated at novel viewpoints, where we evaluate the novel view appearance with photometric indicators and geometry using novel view depth maps. In addition, we employ the Chamfer Distance to evaluate the quality of the overall reconstructed 3D geometry, rather than per-view geometry. For all experiments, each model receives four input views sampled from random positions and is evaluated on ten target viewpoints.

\subsection{Implementation Details}

We randomly choose 5 images from multi-view renderings to form a training batch (4 input views and 1 target view). We train RnG on 256$\times$256 resolution with a patch size of 8. 
The model is trained on 8 A800 GPUs for 40K steps with a total batch size of 96. We use bfloat16 and gradient checkpointing to improve training efficiency. The hyper-parameters to weight terms in the multi-task loss is set to $\lambda_{pmap}=0.2$, $\lambda_c=1$ and $\lambda_\text{p}=0.5$. The uncertainty re-weight term $\alpha=0.2$. Please find more training details in the supplementary. 

\begin{figure*}
\begin{minipage}[t]{0.68\textwidth}
\small
\centering
\resizebox{\linewidth}{!}{
\begin{tabular}{cccc}
{\includegraphics[width=0.25\linewidth]{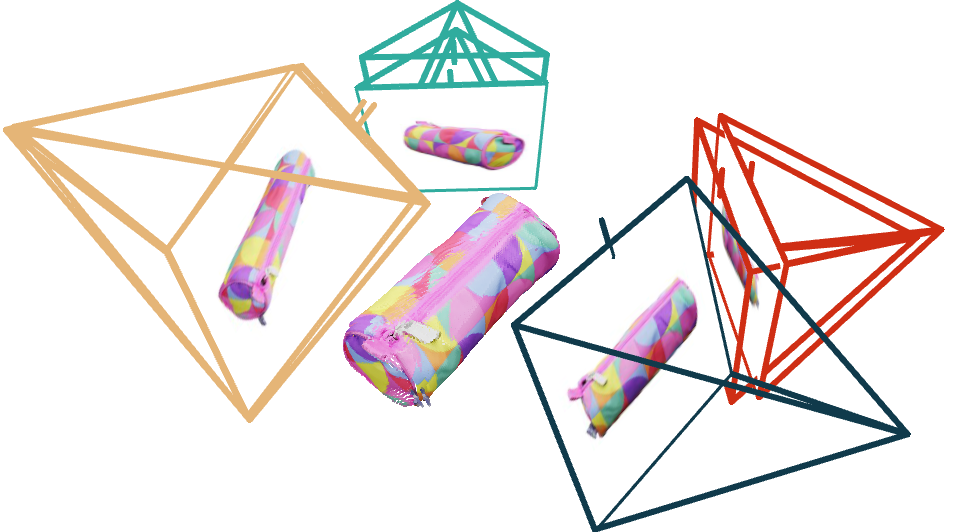}} &
{\includegraphics[width=0.25\linewidth]{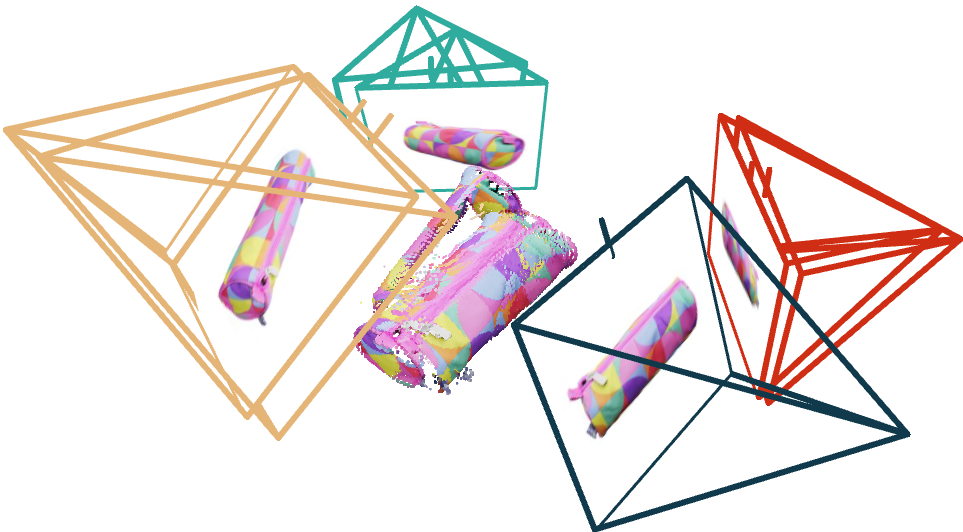}} &
{\includegraphics[width=0.25\linewidth]{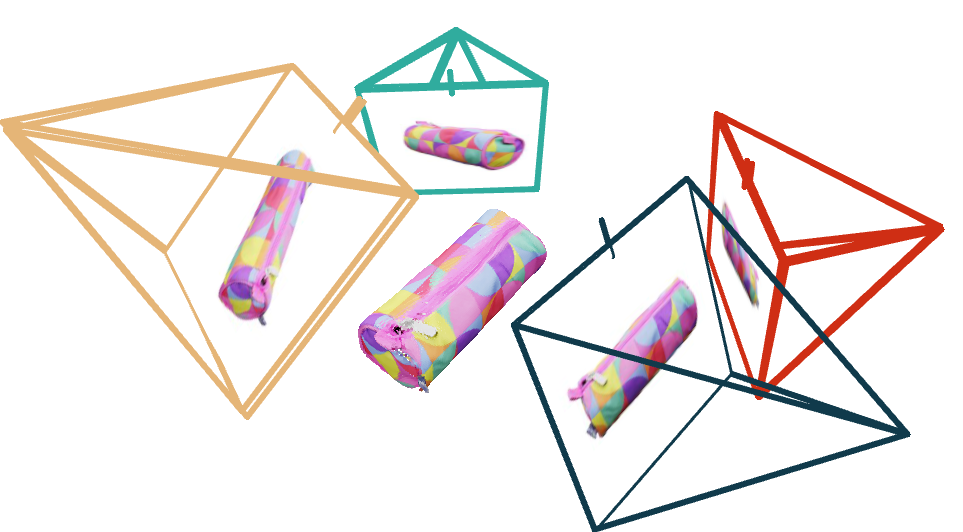}} &
{\includegraphics[width=0.25\linewidth]{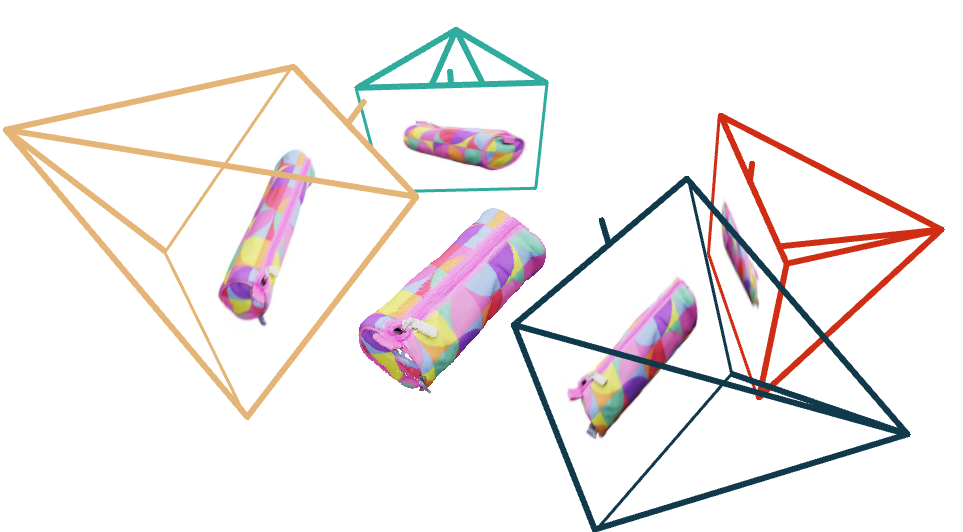}} \\

{\includegraphics[width=0.25\linewidth]{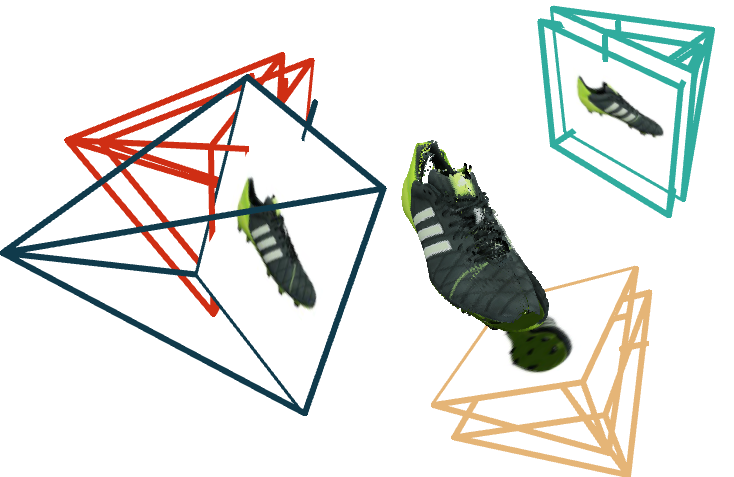}} &
{\includegraphics[width=0.25\linewidth]{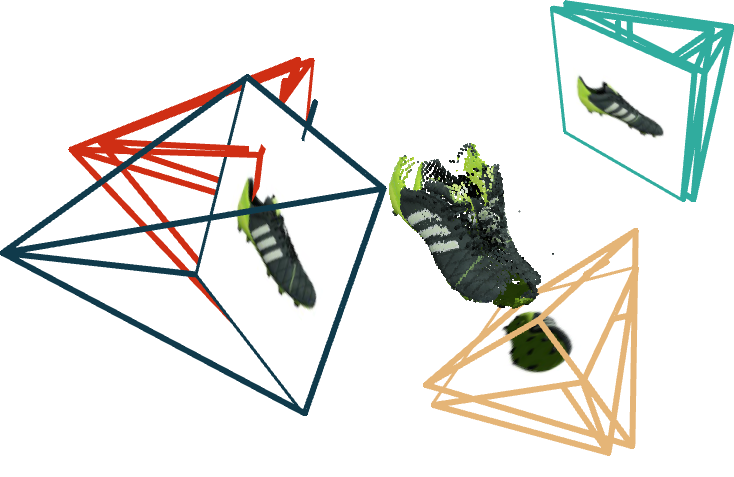}} &
{\includegraphics[width=0.25\linewidth]{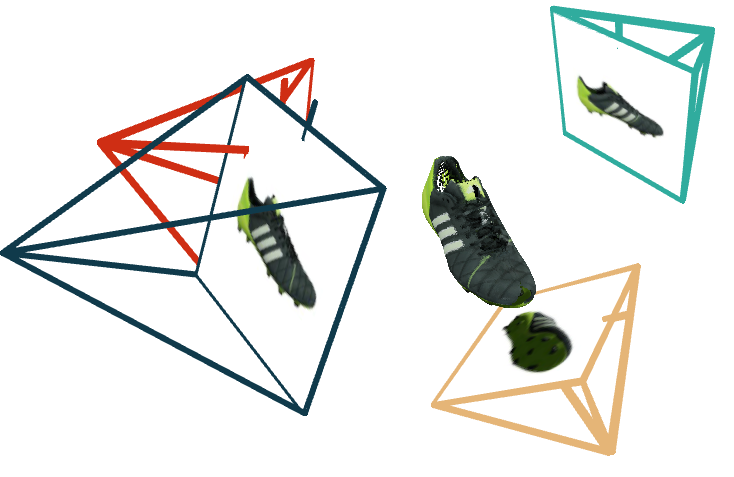}} &
{\includegraphics[width=0.25\linewidth]{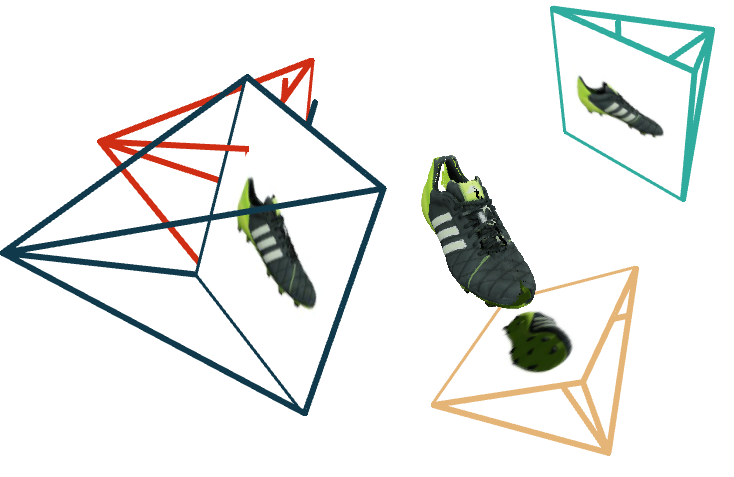}} \\

VGGT~\cite{vggt} & Matrix3D~\cite{matrix3d} & RnG (Ours) & GT \\

\end{tabular}
} 
\caption{\textbf{Camera pose and point cloud visualization.} Reconstructions are normalized to match GT's scale and are aligned to first frame's position (dark blue). The estimated camera pose from RnG highly aligns with the ground truth. Our back-projected point cloud from source views does not suffer from layering artifacts, presenting accurate object structures. }
\label{fig:main_pose}
\end{minipage}
\hfill
\begin{minipage}[t]{0.30\textwidth}
\resizebox{\linewidth}{!}{
\small
\centering
\setlength{\tabcolsep}{1pt}
\begin{tabular}{ccc}
\includegraphics[width=0.33\linewidth, trim={160 100 100 200},clip]{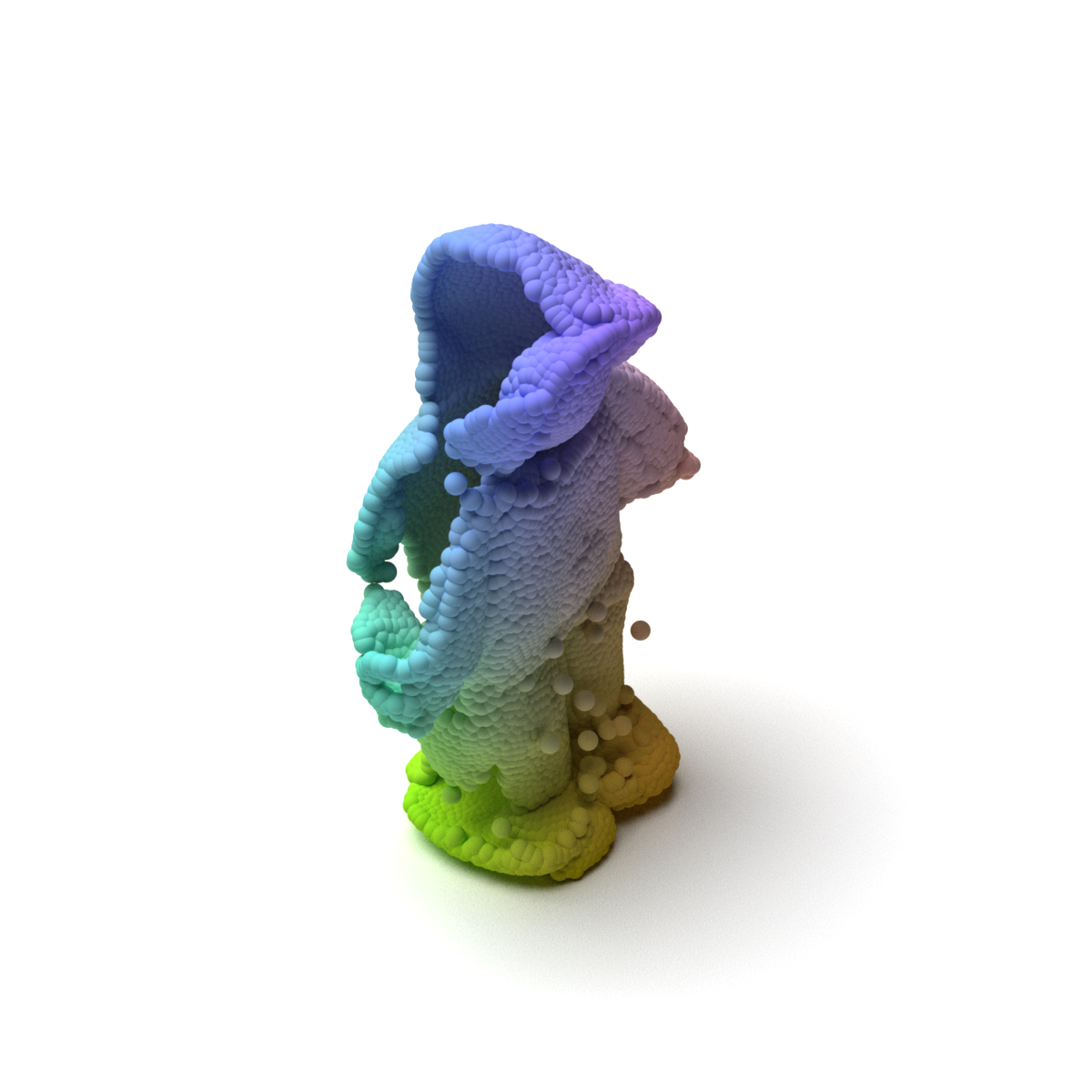} &
\includegraphics[width=0.33\linewidth, trim={160 100 100 200},clip]{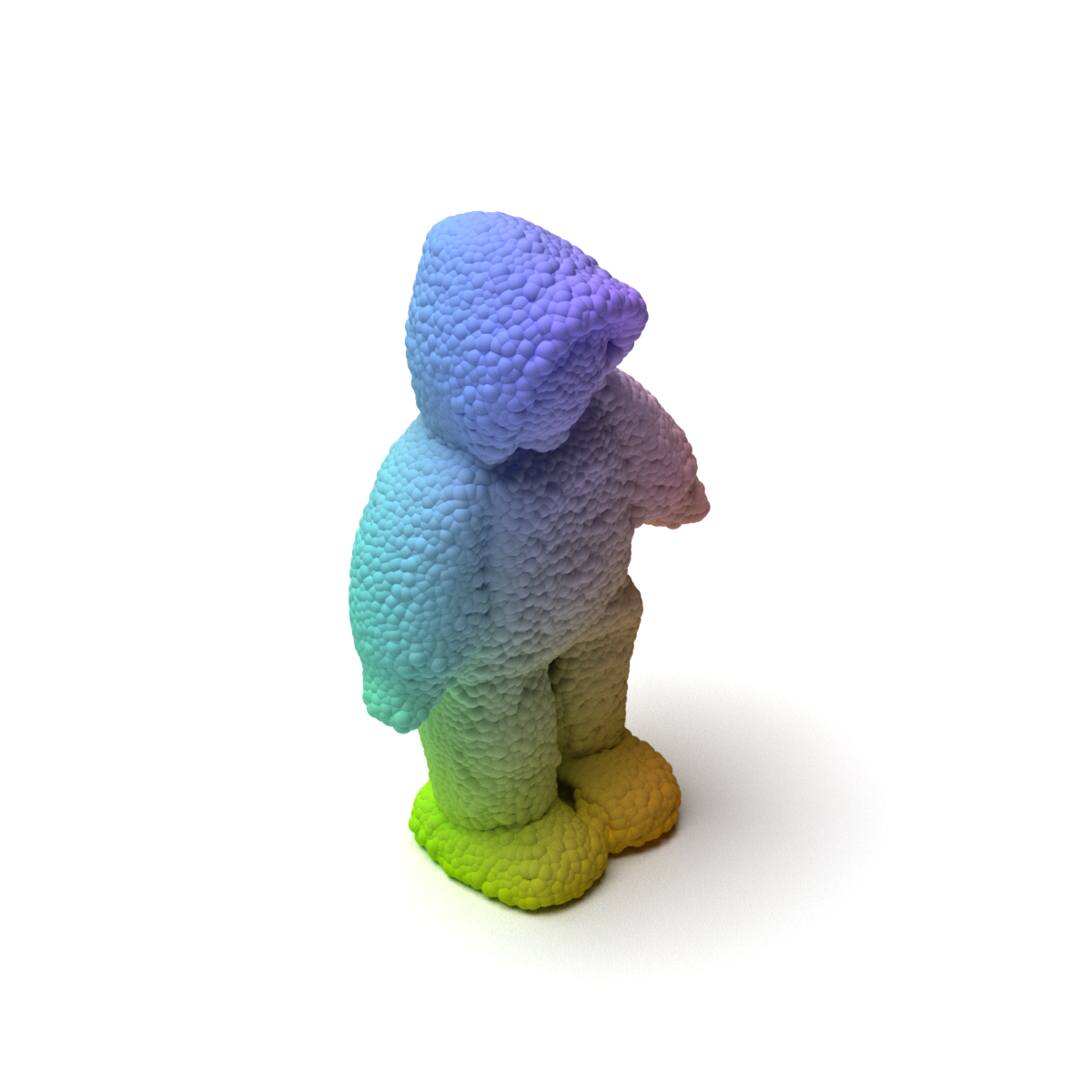} &
\includegraphics[width=0.33\linewidth, trim={160 100 100 200},clip]{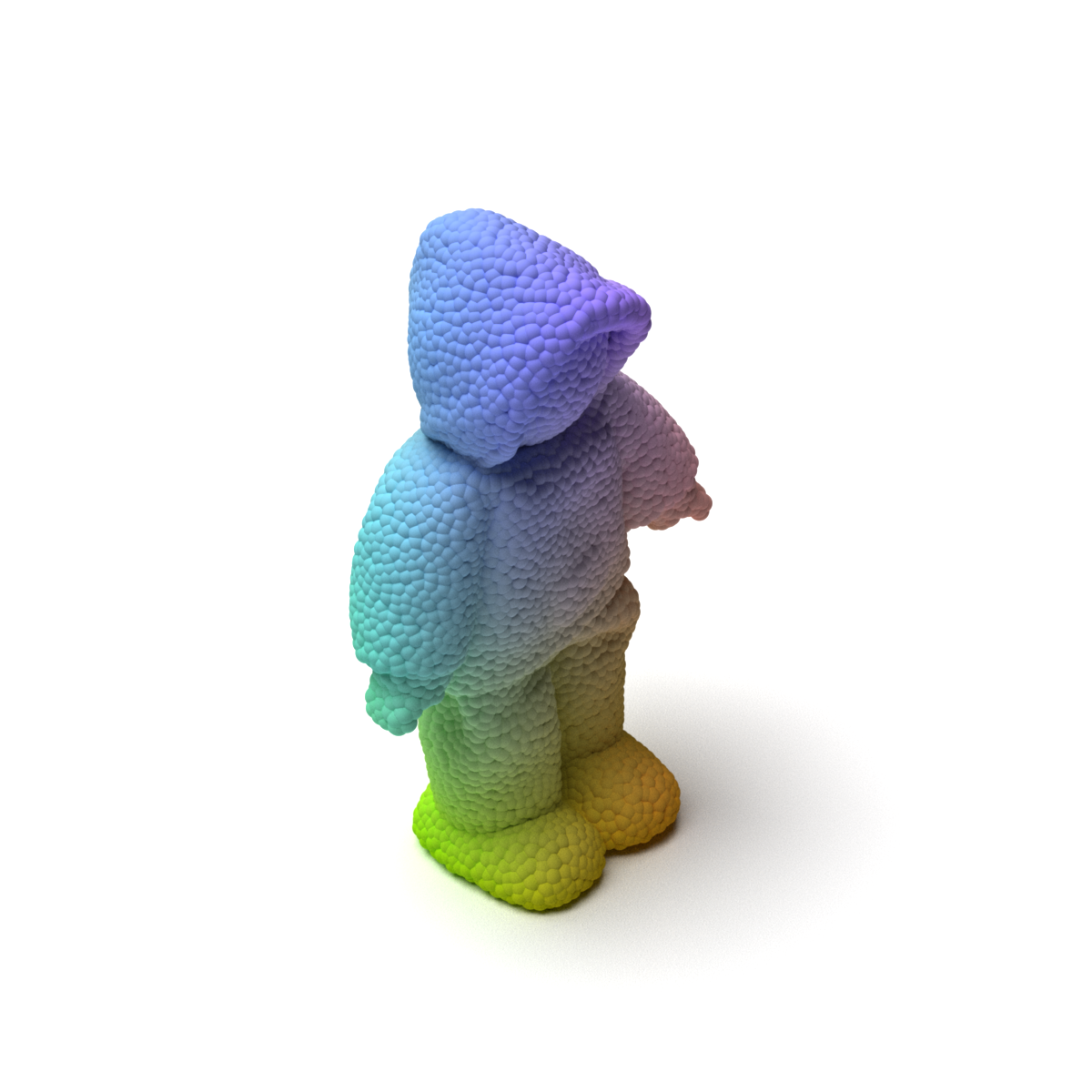} \\

\includegraphics[width=0.33\linewidth, trim={100 100 160 200},clip]{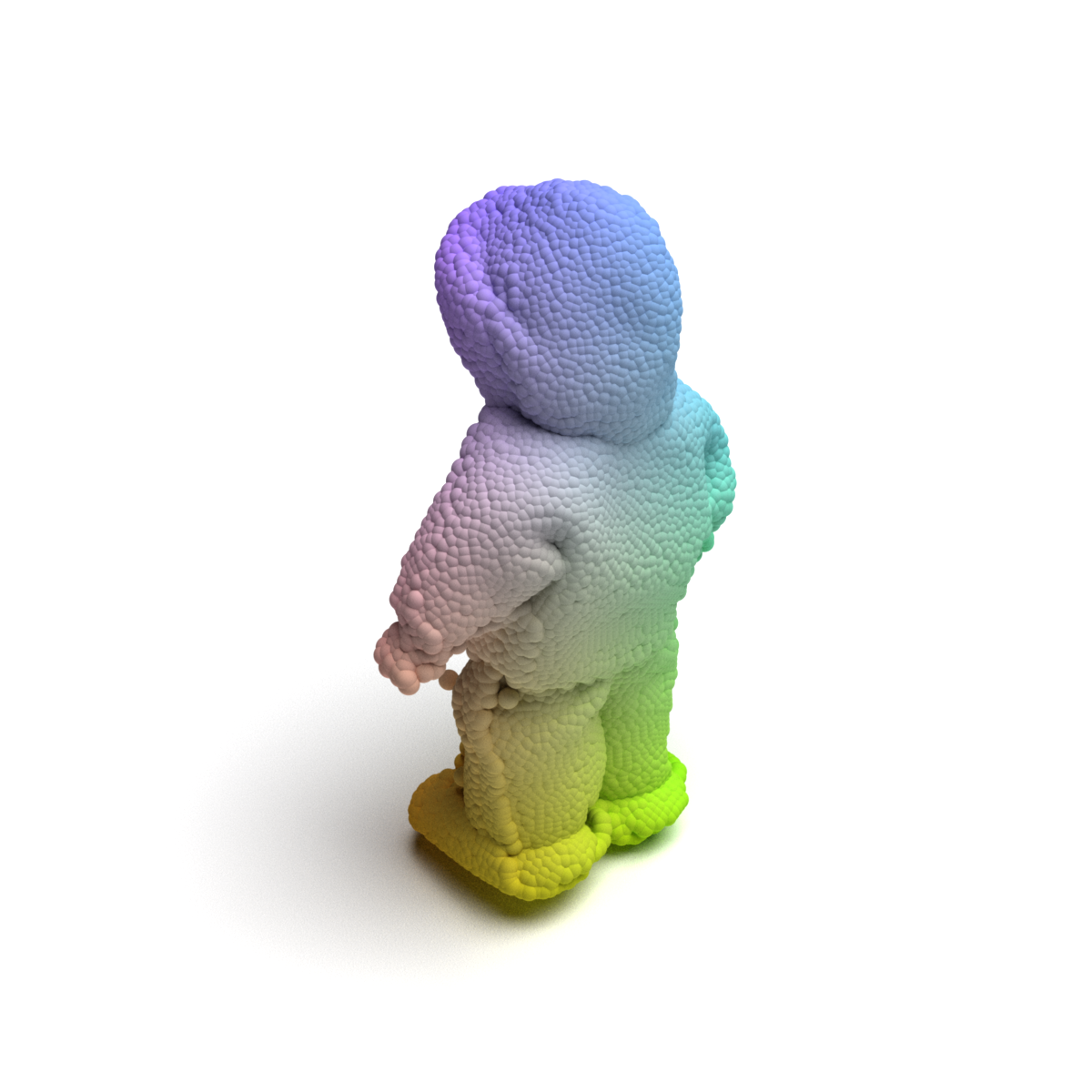} &
\includegraphics[width=0.33\linewidth, trim={100 100 160 200},clip]{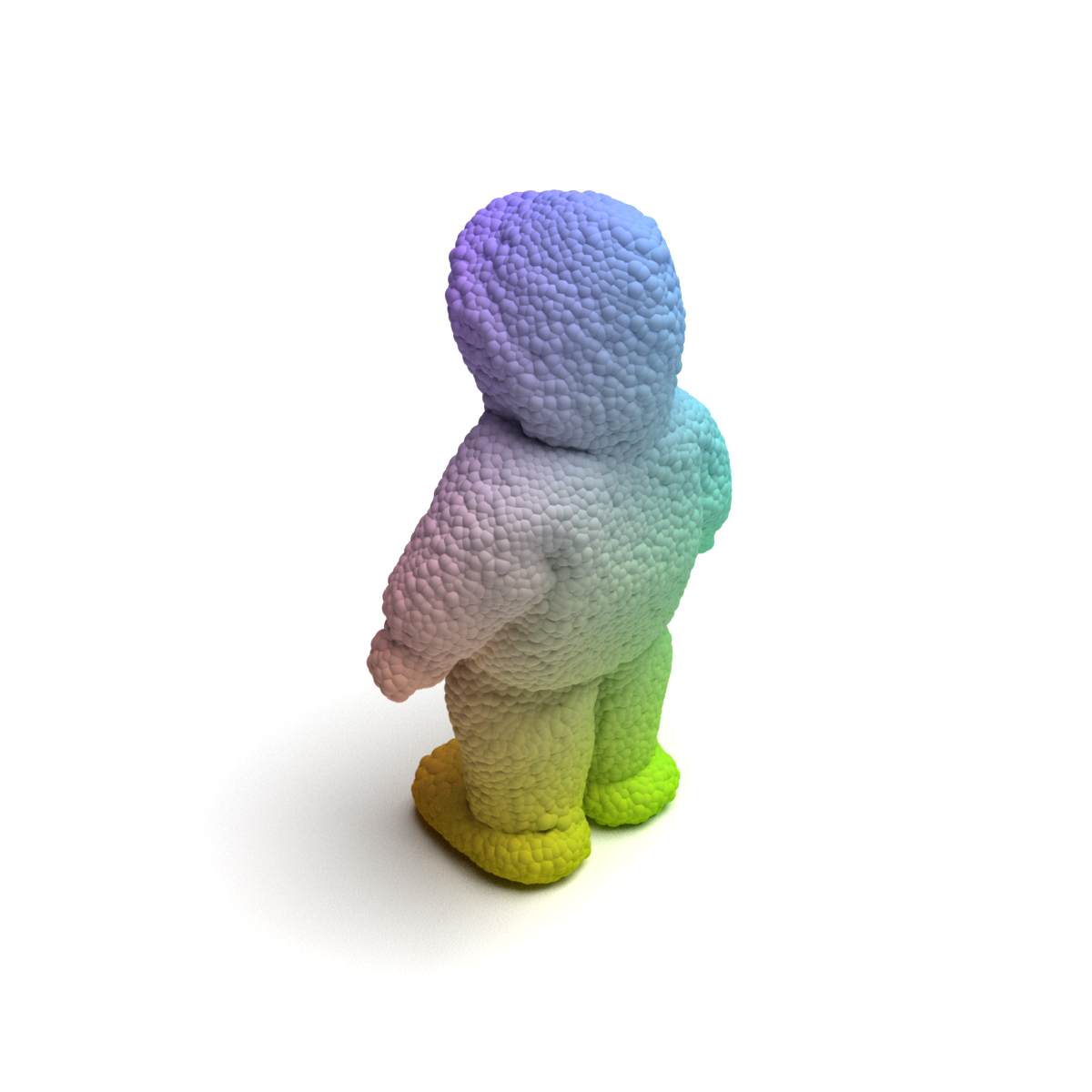} &
\includegraphics[width=0.33\linewidth, trim={100 100 160 200},clip]{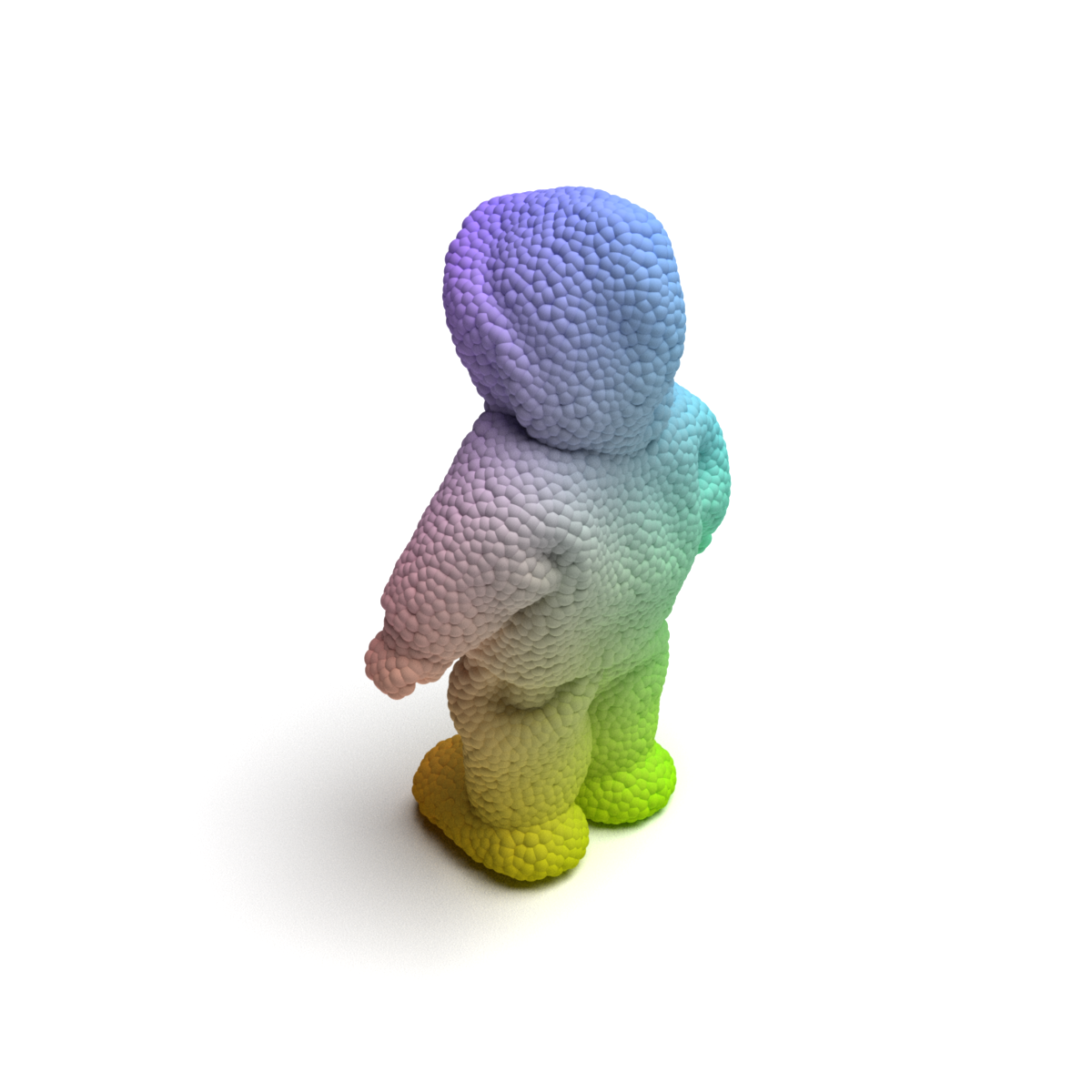} \\

VGGT\cite{vggt} &
RnG (ours) & GT \\

\end{tabular}
} 
\caption{\textbf{3D structures from 4 views.} 
VGGT reconstructs observed region with layering artifacts.
RnG can accumulate complete 3D from multiple view queries.}
\label{fig:pcd_compare}
\end{minipage}

\end{figure*}

\subsection{Experimental Results}
\subsubsection{Comparison with Existing Methods}
We compare RnG against state-of-the-art reconstruction and novel view synthesis methods. In Tab. \ref{tab:main_comparison}, we evaluate our approach on both source-view reconstruction metrics and novel-view generation metrics. Below, we provide detailed analyses for each part.

\noindent \textbf{Input view Reconstruction.} 
We first evaluate the reconstruction capability of each method, covering input-view camera pose estimation and depth prediction accuracy. As shown in~\autoref{tab:main_comparison}, compared with the unified model Matrix3D~\cite{matrix3d}, RnG achieves superior performance in both tasks. 
Compared with the state-of-the-art 3D reconstruction method VGGT~\cite{vggt}, RnG outperforms by a large margin in camera pose and depth accuracy.

The reconstructed point clouds of input view are visualized in~\autoref{fig:main_pose}. The multi-view back-projected point clouds of RnG are highly consistent in 3D.

\noindent \textbf{Novel view appearance generation.}
Next, we evaluate the novel-view appearance generation ability of RnG. Despite being an unposed method, RnG achieves comparable performance to the best pose-required model, LVSM \cite{lvsm}, in novel-view synthesis. Since LVSM cannot directly handle unposed rendering, we use VGGT~\cite{vggt} to obtain pose estimation results, and then feed these results into LVSM~\cite{lvsm} for novel view synthesis, denoted as `VGGT + LVSM'. Among all unposed pipelines, RnG delivers significantly better photo-realism and visual consistency.

We also visualize novel view synthesis results in~\autoref{fig:main_visual}. LGM~\cite{lgm} reconstruct 3D Gaussians from source images, so when observations cannot cover the 3D object, novel views are also incomplete; the multi-view stacked Gaussians also suffer from layering artifacts. Matrix3D~\cite{matrix3d} delivers `eye candy' novel view images, but often fails to keep 3D consistency of the object and may hallucinate non-existing details. When using LVSM~\cite{lvsm} with estimated camera poses from VGGT~\cite{vggt}, the resulting objects are often distorted. 

\noindent \textbf{Novel view geometry generation.}
Finally, we assess the novel-view geometry generation capability of RnG, represented as depth maps. While reconstruction methods such as VGGT \cite{vggt} can only recover 3D structures from source viewpoints, Matrix3D \cite{matrix3d} and RnG can generate depth from arbitrary novel views. Notably, the depth error of RnG is an order of magnitude lower than unposed or even pose-required version of Matrix3D. 

Furthermore, our method produces high-quality multi-view fused point clouds. As shown by the Chamfer Distance reported in~\autoref{tab:main_comparison}, RnG achieves state-of-the-art performance in overall 3D geometry reconstruction. This result further verifies that the point clouds generated by RnG are highly consistent across different viewpoints. \autoref{fig:pcd_compare} shows clear advantages of generating unseen geometry. The reconstructed 3D structure of VGGT~\cite{vggt} is incomplete (only observed regions can be recovered) and presents layering artifacts. In contrast, RnG generates plausible geometry for both source and target viewpoints, resulting in a more complete and structurally consistent 3D reconstruction overall.

In summary, RnG achieves state-of-the-art performance across multiple 3D reconstruction and novel-view generation metrics, including camera pose estimation, source and novel-view depth prediction, reconstructed 3D geometry and RGB rendering. Remarkably, it can even surpasses specialized models designed for individual tasks by a significant margin.

\subsubsection{Generalize to Arbitrary Input Views}
\label{sec:generalize_to_arbitrary_input_views}

\begin{figure}[b]
\setlength{\tabcolsep}{0pt}
\begin{tabular}{c c}
\resizebox{0.33\linewidth}{!}{
\begin{tabular}{c}
\includegraphics[width=\linewidth]{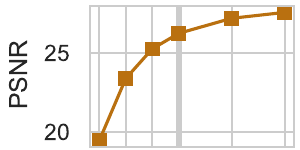} \\
\includegraphics[width=\linewidth]{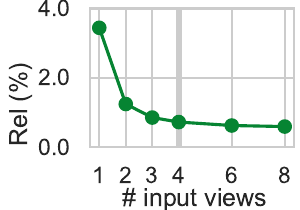} \\
\end{tabular}
}
&
\resizebox{0.65\linewidth}{!}{
\setlength{\tabcolsep}{1pt}
\begin{tabular}{cc | cc c}
\includegraphics[width=0.15\linewidth, trim={60 40 60 30},clip]{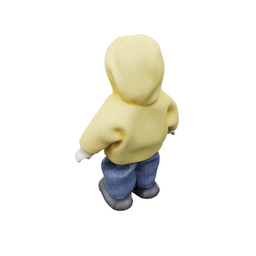} &
\includegraphics[width=0.15\linewidth, trim={60 40 60 30},clip]{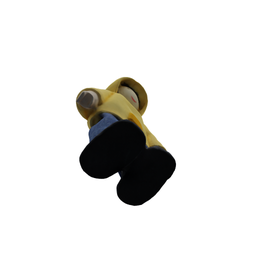} &

\includegraphics[width=0.15\linewidth, trim={70 50 70 40},clip]{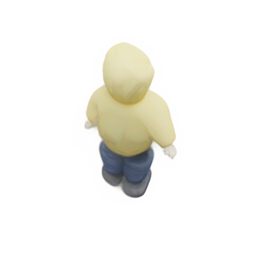} &
\includegraphics[width=0.15\linewidth, trim={70 50 70 40},clip]{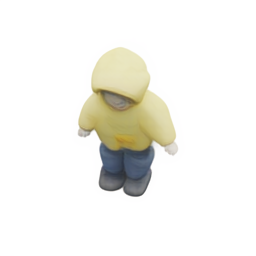} &
\includegraphics[width=0.15\linewidth, trim={70 50 70 40},clip]{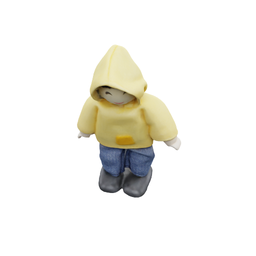} \\ 

\includegraphics[width=0.15\linewidth, trim={30 30 30 30},clip]{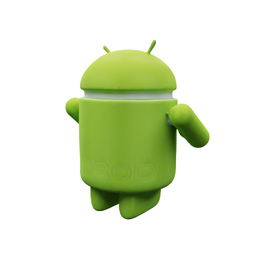} &
\includegraphics[width=0.15\linewidth, trim={30 30 30 30},clip]{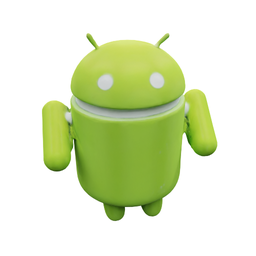} &

\includegraphics[width=0.15\linewidth, trim={40 30 40 40},clip]{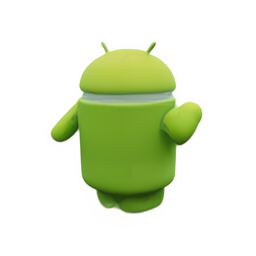} &
\includegraphics[width=0.15\linewidth, trim={40 30 40 40},clip]{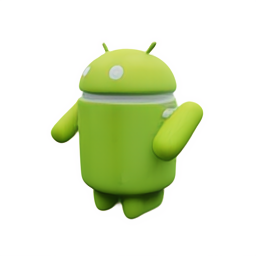} &
\includegraphics[width=0.15\linewidth, trim={40 30 40 40},clip]{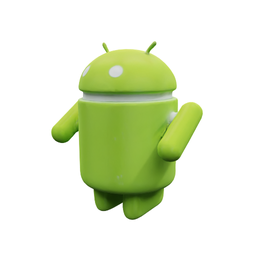} \\
 (a) & (b) & use (a) & use (a+b) & GT \\
\multicolumn{2}{c|}{source views} &
\multicolumn{3}{c}{novel view} \\
\end{tabular}
} 
\end{tabular}
\caption{\textbf{Generalize to other number of input views.} Although our model is not trained to handle other number of input views, it still shows strong generalization ability to other number of source images. RnG even works reasonably when given only 1 image.}
\label{fig:ab_num_view}
\end{figure}

Although RnG is trained with a fixed number of input views, it can directly generalize to an arbitrary number of input views. As shown in~\autoref{fig:ab_num_view}, both the target-view depth and image synthesis quality improves when denser source views are provided. Notably, RnG still achieves reasonable results even under fewer input-view settings.

We also provide visualizations in~\autoref{fig:ab_num_view}. 
When the number of source images increases and more observations become available, the synthesized target images exhibit richer details and higher visual fidelity. Moreover, when the object exhibits symmetric structures, RnG can generate decent novel views even from a single source image.
More experimental results are provided in the supplementary material.

\begin{table*}[t]
\centering
\caption{\textbf{Ablation studies.} We studies the training efficiency of RnG by comparing with LVSM and effectiveness of the model architecture.}
\setlength{\tabcolsep}{4pt}
\resizebox{0.95\linewidth}{!}{
\begin{tabular}{l | ccc | cc | cc | ccc} \toprule
 & \multicolumn{5}{c|}{\textit{Reconstruction}} & \multicolumn{5}{c}{\textit{Generation}} \\
 
 & \multicolumn{3}{c|}{Pose} &
   \multicolumn{2}{c|}{Source View Depth} & 
   \multicolumn{2}{c|}{Novel View Depth} & 
   \multicolumn{3}{c}{Novel View Synthesis} \\

\multicolumn{1}{c|}{Model} & RA@5$\uparrow$ & RT@5$\uparrow$ & AUC@30$\uparrow$ &
     Rel$\downarrow$ & a1$\uparrow$ & Rel$\downarrow$ & a1$\uparrow$ &
     PSNR$\uparrow$ & SSIM$\uparrow$ & LPIPS$\downarrow$ \\ \midrule

LVSM-100K &\multicolumn{7}{c|}{
\multirow{2}{*}{\textit{not applicable}}
}& 27.522 & 0.902 & 0.0895 \\

LVSM-40K  &\multicolumn{7}{c|}{}& 24.619 & 0.864 & 0.1544 \\

Ours-40K & 85.146 & 86.019 & 86.942 & 
            0.584 & 99.929 &  0.717 & 99.850 & 
           26.276 &  0.891 & 0.0975 \\  \midrule

Ours-15K-scratch  &~~8.252 & 17.184 & 33.654 &
            2.331 & 99.475 &  2.331 & 99.364 &
           20.784 &  0.819 & 0.2035 \\
           
Ours-15K & 81.650 & 82.233 & 82.631 &
            0.742 & 99.884 &  0.882 & 99.788 &
           24.856 &  0.873 & 0.1243 \\

Ours-15K-w/o cam  & \multicolumn{3}{c|}{\textit{not applicable}} &
            0.739 & 99.890 &  0.878 & 99.789 &
           24.850 &  0.873 & 0.1235 \\

Ours-15K-FullAttn & 82.718 & 81.456 & 82.686 &
            0.727 & 99.906 &  0.892 & 99.792 &
           24.861 &  0.874 & 0.1193 \\
         \bottomrule
\end{tabular}
} 
\label{tab:ab}
\end{table*}

\subsection{Ablation Studies}

We further conduct ablation studies to demonstrate the effectiveness of our model architecture. The results are presented in~\autoref{tab:ab}. 

\noindent\textbf{Model Scaling.}
In~\autoref{tab:ab}, `Ours-15K' models are trained on the LVIS subset containing 44.5K objects, whereas the main model, `ours-40k', is trained on the full dataset (113.5K objects) with additional training iterations, resulting in superior performance. These results highlight the scalability and strong scaling potential of RnG.

\noindent\textbf{Training Efficiency.}
We train two variants of LVSM. LVSM-100K is trained for 100K iterations and reflects the best achievable performance of LVSM. LVSM-40K is trained for 40K steps, the same as RnG to provide a fair comparison. As displayed in~\autoref{tab:ab}, Ours-40K outperforms LVSM-40K on novel view synthesis metrics, demonstrating that our proposed approach has the potential to surpass state-of-the-art pose-conditioned novel view synthesis methods when trained for more iterations. Moreover, Ours-15K already exceeds the performance of LVSM-40K despite being trained on a much smaller dataset, 
highlighting the effectiveness of using 3D reconstruction prior to generate novel view appearance and geometry.

\noindent\textbf{Reconstruction Prior.}
We also train RnG without the pretrained VGGT~\cite{vggt} weights, denoted as Ours-15K-scratch. There is a significant performance drop compared to the model initialized with pretrained weights. This demonstrates that incorporating reconstruction prior knowledge can effectively enhance our generation tasks. Additionally, we train a variant without the camera head (Ours-15K-w/o cam), showing that removing camera pose supervision does not negatively impact novel view synthesis performance. 
This means that in our multi-task learning paradigm, learning to reconstruct camera poses and generate novel view do not conflict with each other. With the estimated camera pose as a free gift, we can easily map the texture of source images to get finer details.

\noindent\textbf{Reconstruction-Guided Causal Attention.}
Our core design choice, the reconstruction-guided causal attention, is motivated by decoupling reconstruction and generation. As detailed in Sec~\ref{sec:KV-Cache}, this design enables a two-stage inference process that utilizes a KV-Cache, significantly improving efficiency (reducing inference time from 213ms to 85ms on an A800 GPU; in contrast, Matrix3D~\cite{matrix3d} relies on a denoising diffusion process and requires 27s to synthesize a single novel view.). 
Detailed statistics are presented in~\autoref{tab:efficiency}.
To validate that this efficient architecture does not come at the cost of accuracy, we ablate it by replacing our causal mask with full bidirectional attention ({Ours-15K-FullAttn}). As shown in~\autoref{tab:ab}, this change yields a negligible performance difference. This confirms our causal design achieves its significant architectural advantages: enforcing a principled information flow and enabling the KV-Cache without compromising performance.

\begin{table}[t]
\vspace{-0.1in}
\caption{\textbf{Efficiency comparison} of models with and without KV-Cache for inferring a single novel-view appearance and geometry.}
\vspace{-0.12in}
\centering
\small
\begin{tabular}{l|ccc} \toprule
    & peak memory & FLOPS & time \\ \midrule
Matrix3D \cite{matrix3d} & 11.31 G & 2228 T & 27 s \\ \midrule
RnG w/o KV-Cache      & 8.12 G &  12.26 T &  213 ms \\
RnG w/ ~~KV-Cache & 7.10 G & ~~2.29 T & ~~85 ms \\ \bottomrule
\end{tabular}
\label{tab:efficiency}
\end{table}

\section{Limitations}
\label{sec:limitations}

Although RnG demonstrates leading performance in both reconstructing the source view structure and generating novel view geometry and appearance, it also has limitations.

\noindent \textbf{Lack of details.} Unlike diffusion-based image generators that produce eye-catching images with very rich details (although they can be wrong), RnG has difficulties in representing intricate textures. This is also an issue in LVSM~\cite{lvsm}. Introducing image-generation pretraining or auto-encoding components might resolve this issue.

\noindent \textbf{World origin definition.} For object-level reconstruction task, an ideal world origin would be the object center. However, in data preparation, all cameras look at the world center, which may confuse the network to regard the intersection of source views' optical axes as the world origin. This also limits model's real-world applications, where multi-view images captured by hand-held devices have to look at a common 3D point to get the best result. Improving the data-processing pipeline can resolve this issue.

\noindent \textbf{Accumulating 3D from multiple viewpoints.} Although our model generates highly consistent point maps from viewpoint queries, it requires to accumulate multi-view observations to obtain a complete 3D. Multi-view geometry integration might also introduce noise and conflicts. 

\section{Conclusion}

In this paper, we introduced RnG (Reconstruction and Generation), a novel feed-forward Transformer that addresses the limitation of partiality in generalizable 3D reconstruction. 
Its reconstruction-guided causal attention mechanism leverages the Transformer's KV-Cache as an implicit, complete 3D representation. This allows RnG to efficiently predict and render high-fidelity novel-view RGBD outputs from arbitrary query poses, effectively transforming implicit structural understanding into explicit geometric realization. 
Consequently, RnG not only accurately reconstructs visible geometry but also successfully generates plausible, coherent unseen geometry and appearance.
Our approach achieves state-of-the-art performance across both reconstruction and novel-view generation benchmarks, while being efficient enough for real-time interactive applications. 
For future work, we aim to extend the RnG framework to handle broader scenarios, including dynamic and non-rigid content, positioning RnG as a general-purpose foundation for unified 3D reconstruction and generation.

\section*{Acknowledgments}
This research was supported in part by the National Natural Science Foundation of China (62525115, U2570208).

{
    \small
    \bibliographystyle{ieeenat_fullname}
    \bibliography{main}
}

\clearpage
\maketitlesupplementary

\section{Training Details}

For a single object, we render RGBD images at 25 viewpoints with different radius. To form a training batch, we randomly sample 7 images from the multi-view renderings. The first 4 images will be used as source views, and the rest 3 are different target views. We then expand the batch size by 3 so that each training batch contains the same 4 source images and 1 target image. 

The training batch size is 6 (objects), we accumulate every 2 forward steps for an increased batch size. The maximum learning rate is set to 6e-4. We use learning-rate warm-up for the first 3000 steps and use the cosine learning-rate decay for the rest of the training process. The single image feature extractor's parameters (the DINO vision transformer~\cite{dinov2} in VGGT~\cite{vggt}) are kept frozen during training.

\section{Recovering complete 3D}

\subsection{Recovering up-direction}
\begin{figure}[h]
    \centering
    \includegraphics[width=0.75\linewidth]{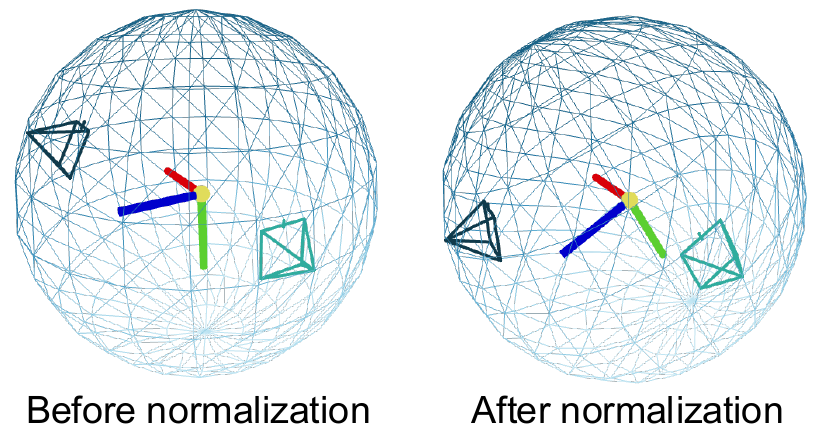}
    \caption{\textbf{The camera normalization process.} The first camera pose will be normalized on a unit sphere and transformed to a fixed position. This introduces additional roll angles to other cameras.}
    \label{fig:supp_cam_norm}
\end{figure}

As illustrated in~\autoref{fig:supp_cam_norm}, the training samples undergo a camera normalization process, where the first camera is scaled and transformed to $\left[I_{3\times3} \left| [0, 0, -1]^\intercal \right. \right]$. This will change the `up direction' of the object in the world coordinate system.
When querying from a novel view, if we directly use pre-defined camera locations, an extra `rolling angle' will be introduced. In other words, the `up-direction' of the world coordinate system does not align with the observed object's up-direction, but with an extra rolling angle. This is out of the training samples' distribution, so after model predicting the camera poses, we have to recover object's up-direction (if possible). 

We assume all input images' up-directions are aligned with the object's up-direction, then the objective is to find a rotation along the $x$-axis (the red axis in~\autoref{fig:supp_cam_norm}) so that all cameras do not have additional `roll angles'. A simple solution is to search for the best angle of elevation that lead to the minimum of all rolling angles.

\subsection{Accumulating views}

\begin{figure}[h]
\centering
\includegraphics[width=\linewidth]{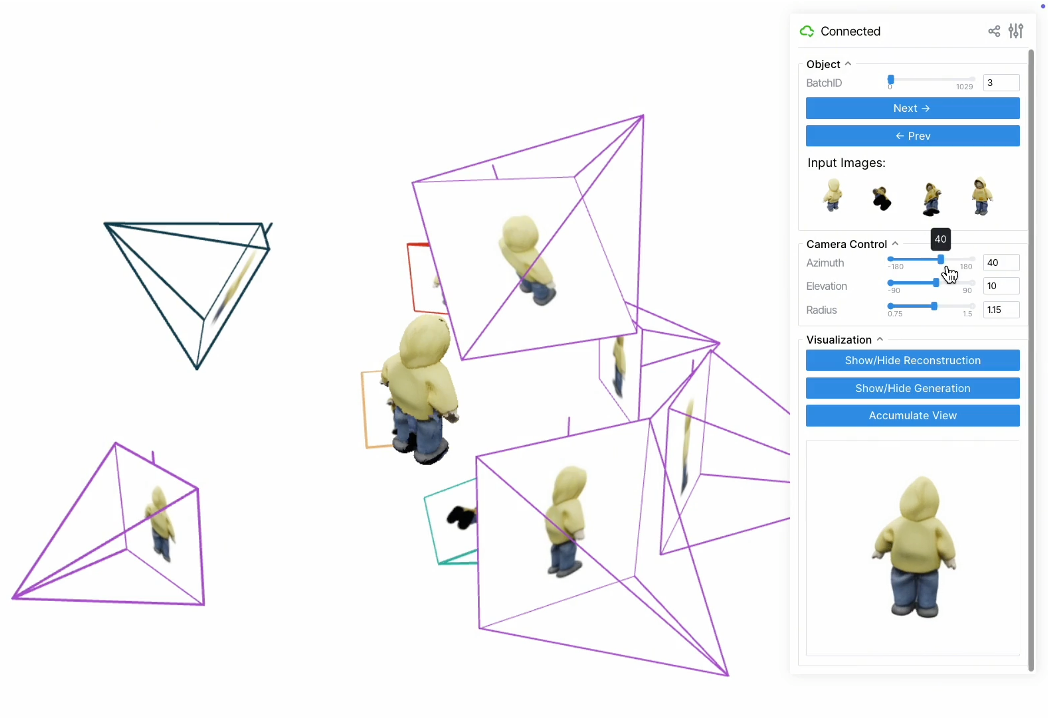}
\caption{The user interface to visualize novel view images and accumulate complete 3D from multiple views.}
\label{fig:supp_ui}
\end{figure}

When extracting complete 3D from the network, we can manually choose viewpoints to back-project the estimated 3D points using our user interface (shown in~\autoref{fig:supp_ui}). We can also uniformly place cameras on a sphere to get novel views. We utilize the estimated confidence map to filter the estimated points to remove flying points around sharp boundaries. The user interface is implemented using Viser~\cite{viser}.

\section{More Experimental Results}

\begin{table*}[t]
\centering
\caption{\textbf{Generalize to other number of input views.} Although our model is not trained to handle other number of input views, it still shows strong generalization ability to other number of source images.}
\setlength{\tabcolsep}{4pt}
\small
\begin{tabular}{c | ccc | cc | cc | ccc} \toprule
 & \multicolumn{5}{c|}{\textit{Reconstruction}} & \multicolumn{5}{c}{\textit{Generation}} \\
 
 \# input & \multicolumn{3}{c|}{Pose} &
   \multicolumn{2}{c|}{Input View Depth} & 
   \multicolumn{2}{c|}{Novel View Depth} & 
   \multicolumn{3}{c}{Novel View Synthesis} \\

 views & RA@5$\uparrow$ & RT@5$\uparrow$ & AUC@30$\uparrow$ &
     Rel$\downarrow$ & a1$\uparrow$ & Rel$\downarrow$ & a1$\uparrow$ &
     PSNR$\uparrow$ & SSIM$\uparrow$ & LPIPS$\downarrow$ \\ \midrule
8 & 85.922 & 86.893 & 87.942 &
     0.535 & 99.927 &  0.593 & 99.875 &
    27.608 &  0.903 & 0.0857 \\

6 & 85.825 & 86.505 & 87.741 &
     0.539 & 99.935 &  0.622 & 99.876 & 
    27.235 &  0.899 & 0.0891 \\

4 & 85.146 & 86.019 & 86.942 & 
     0.584 & 99.929 &  0.717 & 99.850 & 
    26.276 & 0.891  & 0.0975 \\ 
    
3 & 85.243 & 86.311 & 86.136 &
     0.662 & 99.931 &  0.848 & 99.820 &
    25.291 & 0.881  & 0.1063 \\
    
2 & 83.010 & 87.961 & 85.405 &
     0.931 & 99.906 &  1.243 & 99.670 &
    23.403 &  0.860 & 0.1249 \\

1 & \multicolumn{3}{c|}{\textit{not applicable}} &
     3.328 & 99.160 &  3.443 & 98.121 &
    19.452 &  0.811 & 0.1821 \\
\bottomrule
\end{tabular}
\label{tab:supp_input_views}
\end{table*}

\subsection{Generalize to arbitrary input views}
In~\ref{sec:generalize_to_arbitrary_input_views} of the main paper, we demonstrated that the proposed method generalizes to arbitrary input views. In this section, we provide additional details of the experimental setup, as well as more comprehensive quantitative and qualitative results.
Specifically, for each 3D object in the evaluation dataset GSO~\cite{gso_dataset}, we render RGBD frames at a 25 different places. We randomly pick 10 frames for evaluation, the source images are chosen from the rest 15 views. Our main comparison uses 4 random source views. To keep the metrics comparable when given different number of source views, we remove or append to these 4 views and keep the target views fixed. 

The detailed metric results are given in~\autoref{tab:supp_input_views}.  
As shown, both the target-view depth and image synthesis quality improves when denser source views are provided. Notably, RnG still achieves reasonable results even under two input-view settings. For instance, the PSNR of novel view appearance decreases by only 7.5\% compared with our standard training setup using four input views. A similar trend is observed across other metrics, further validating the strong generalization ability of the proposed method.

\begin{figure*}
\centering
\small
\begin{tabular}{c cccc cc}
1 source view&
\begin{tabular}{c}
\includegraphics[width=0.1\linewidth]{figs/num_input/3/input_0.png}
\end{tabular}
& 
  & 
  & 
  & 
\begin{tabular}{c}
\includegraphics[width=0.1\linewidth]{figs/num_input/3/s1_pred_0.png}
\end{tabular}
& 
\begin{tabular}{c}
\includegraphics[width=0.1\linewidth]{figs/num_input/3/gt_0.png}
\end{tabular}
\\

2 source views &
\begin{tabular}{c}
\includegraphics[width=0.1\linewidth]{figs/num_input/3/input_0.png} 
\end{tabular}
& 
\begin{tabular}{c}
\includegraphics[width=0.1\linewidth]{figs/num_input/3/input_1.png} 
\end{tabular}
 & 
 &
 & 
\begin{tabular}{c}
\includegraphics[width=0.1\linewidth]{figs/num_input/3/s2_pred_0.png} 
\end{tabular}
& 
\begin{tabular}{c}
\includegraphics[width=0.1\linewidth]{figs/num_input/3/gt_0.png} 
\end{tabular}
\\

3 source views & 
\begin{tabular}{c}
\includegraphics[width=0.1\linewidth]{figs/num_input/3/input_0.png} 
\end{tabular}
& 
\begin{tabular}{c}
\includegraphics[width=0.1\linewidth]{figs/num_input/3/input_1.png} 
\end{tabular}
& 
\begin{tabular}{c}
\includegraphics[width=0.1\linewidth]{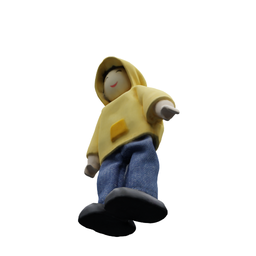} \end{tabular}
& &
\begin{tabular}{c}
\includegraphics[width=0.1\linewidth]{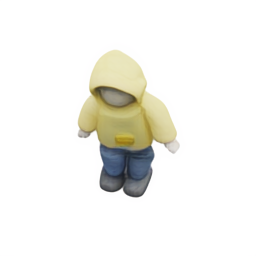} 
\end{tabular}
& 
\begin{tabular}{c}
\includegraphics[width=0.1\linewidth]{figs/num_input/3/gt_0.png} \end{tabular}
\\

4 source views & 
\begin{tabular}{c}
\includegraphics[width=0.1\linewidth]{figs/num_input/3/input_0.png} \end{tabular}
& 
\begin{tabular}{c}
\includegraphics[width=0.1\linewidth]{figs/num_input/3/input_1.png} \end{tabular}
& 
\begin{tabular}{c}
\includegraphics[width=0.1\linewidth]{figs/num_input/3/input_2.png} \end{tabular}
& 
\begin{tabular}{c}
\includegraphics[width=0.1\linewidth]{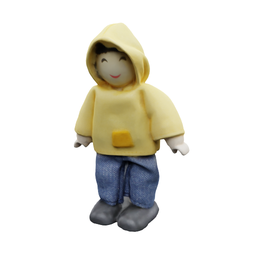} \end{tabular}
& 
\begin{tabular}{c}
\includegraphics[width=0.1\linewidth]{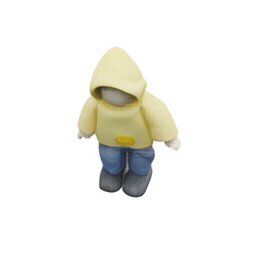} \end{tabular}
& 
\begin{tabular}{c}
\includegraphics[width=0.1\linewidth]{figs/num_input/3/gt_0.png} \end{tabular}
\\

 & source 1 & source 2 & source 3 & source 4 & target prediction & target GT \\

\end{tabular}
\caption{The evaluation protocol when models are given different number of input views. Each row represents an input configuration.}
\label{fig:supp_eval_protocol}
\end{figure*}

\begin{figure*}
\setlength{\tabcolsep}{1pt}
\small
\begin{tabular}{c|c}
\includegraphics[width=0.15\linewidth]{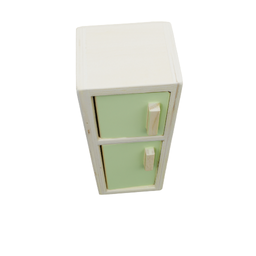} &
\includegraphics[width=0.81\linewidth]{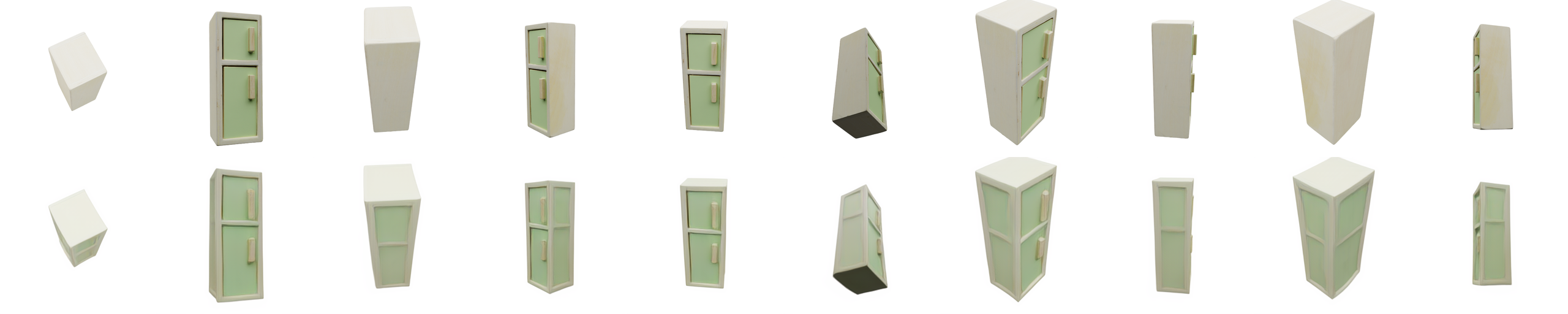} \\
\includegraphics[width=0.15\linewidth]{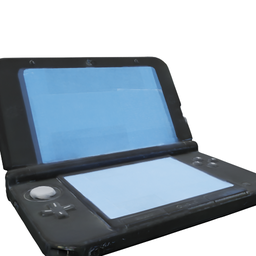} &
\includegraphics[width=0.81\linewidth]{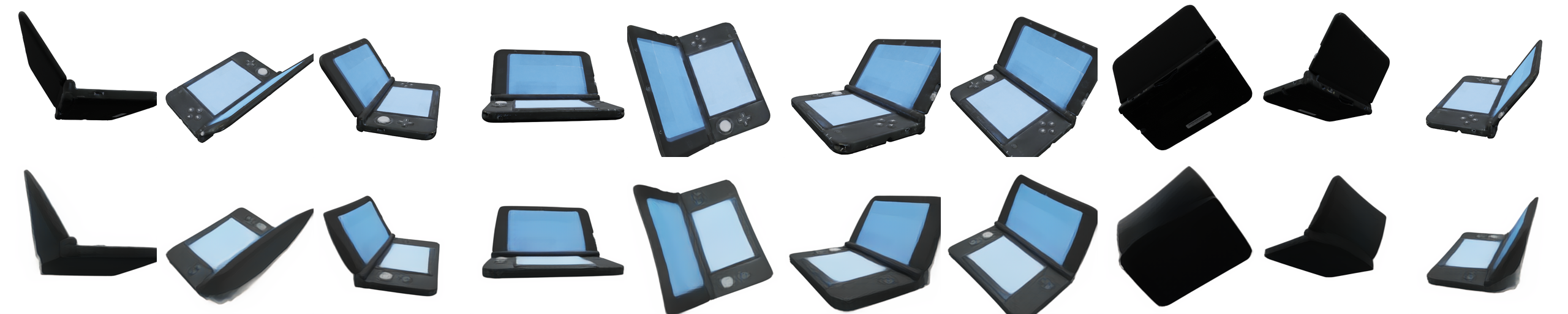} \\
\includegraphics[width=0.15\linewidth]{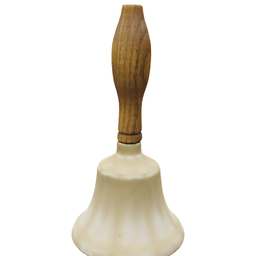} &
\includegraphics[width=0.81\linewidth]{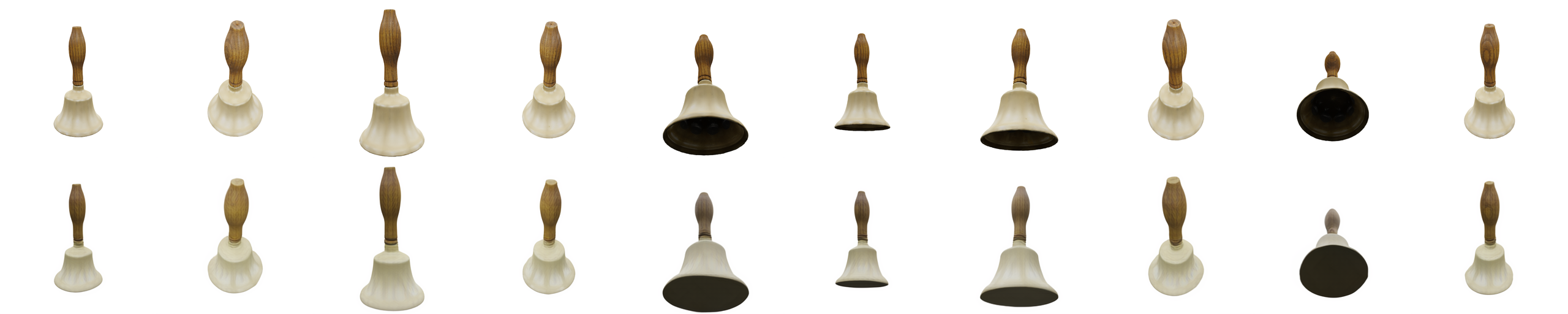} \\
\includegraphics[width=0.15\linewidth]{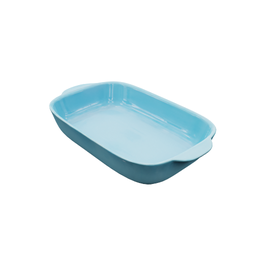} &
\includegraphics[width=0.81\linewidth]{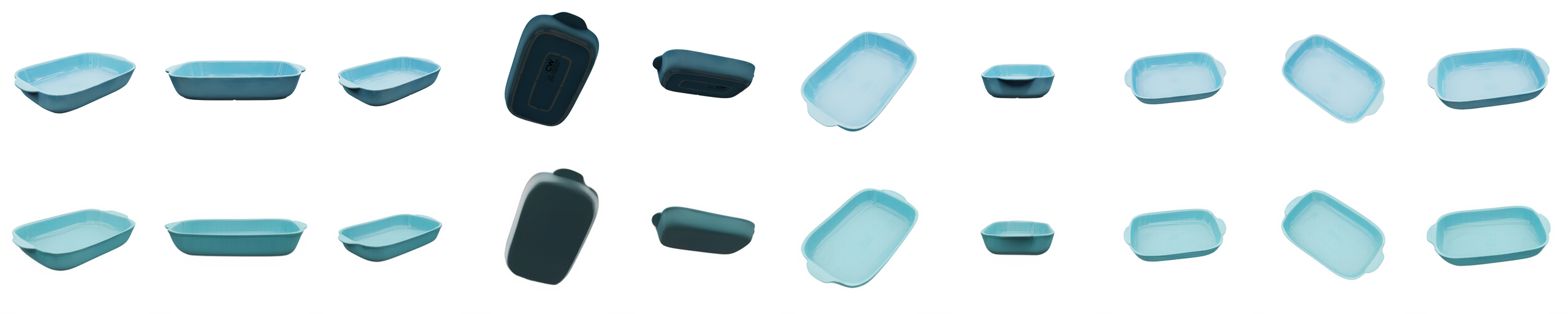} \\
source view & first rows: GT ~~~~~~~~ second rows: model prediction \\
\end{tabular}
\caption{RnG performs well even when only a single image is given.}
\label{fig:supp_one_input_view_good}
\end{figure*}

\begin{figure*}
\setlength{\tabcolsep}{1pt}
\small
\begin{tabular}{c|c}
\includegraphics[width=0.15\linewidth]{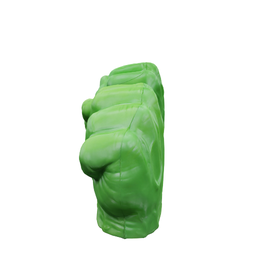} &
\includegraphics[width=0.81\linewidth]{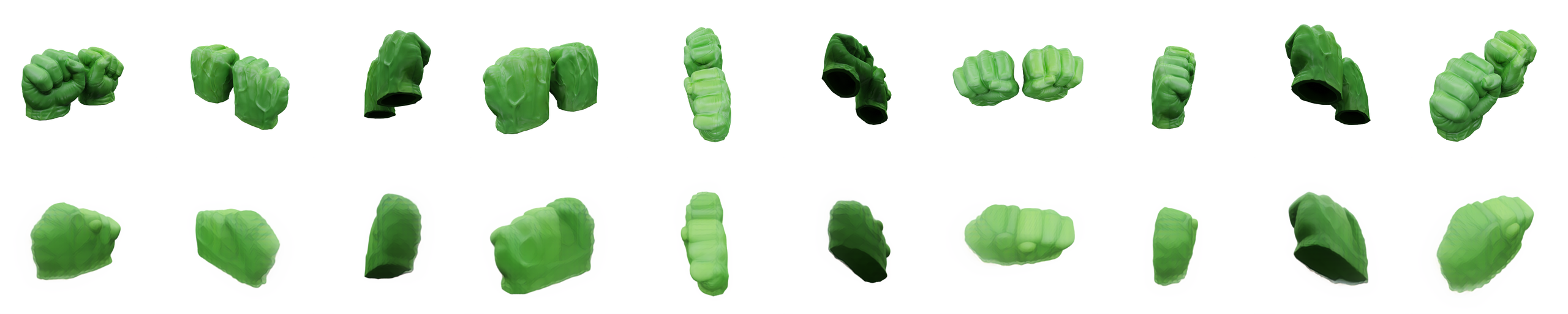} \\
\includegraphics[width=0.15\linewidth]{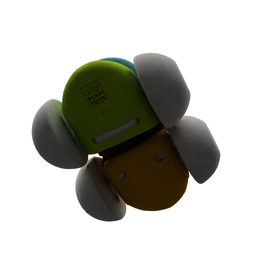} &
\includegraphics[width=0.81\linewidth]{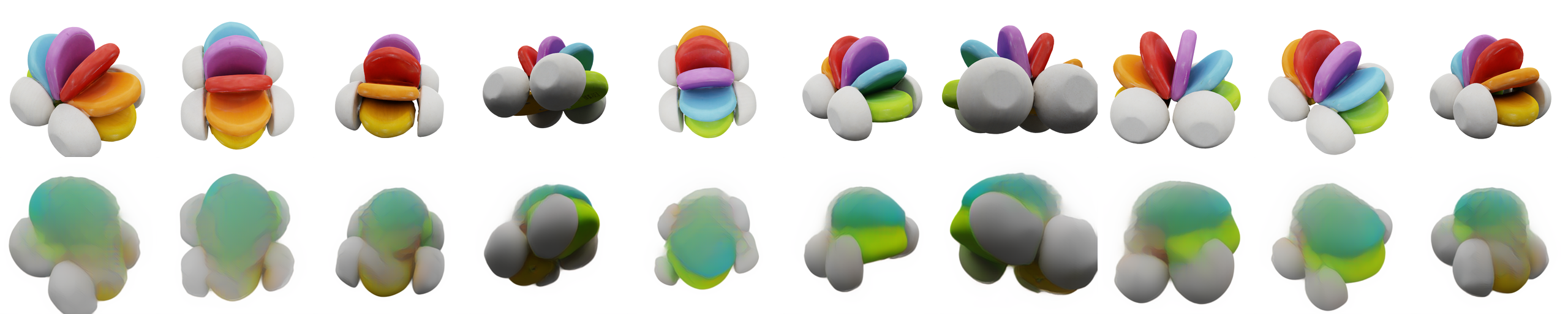} \\
\includegraphics[width=0.15\linewidth]{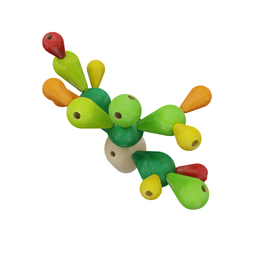} &
\includegraphics[width=0.81\linewidth]{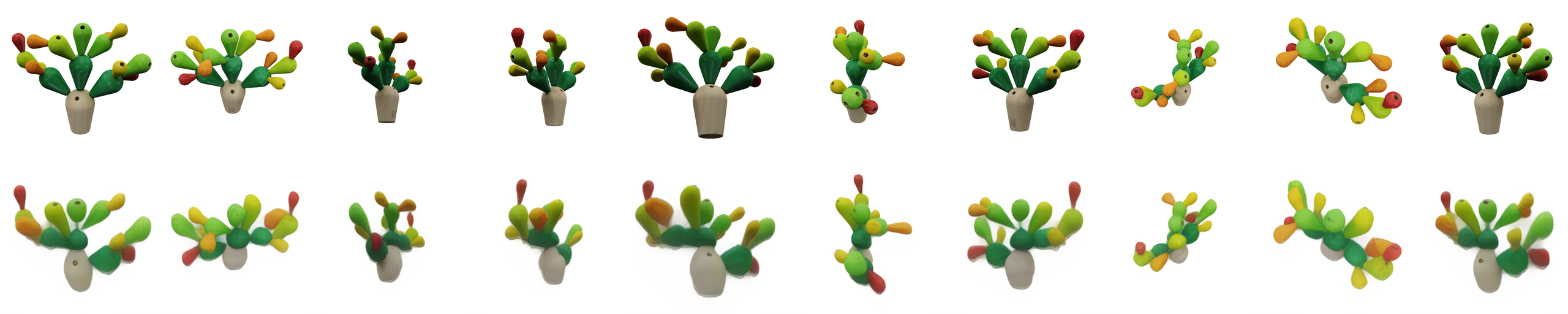} \\
\includegraphics[width=0.15\linewidth]{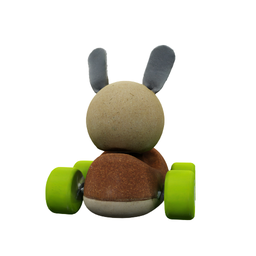} &
\includegraphics[width=0.81\linewidth]{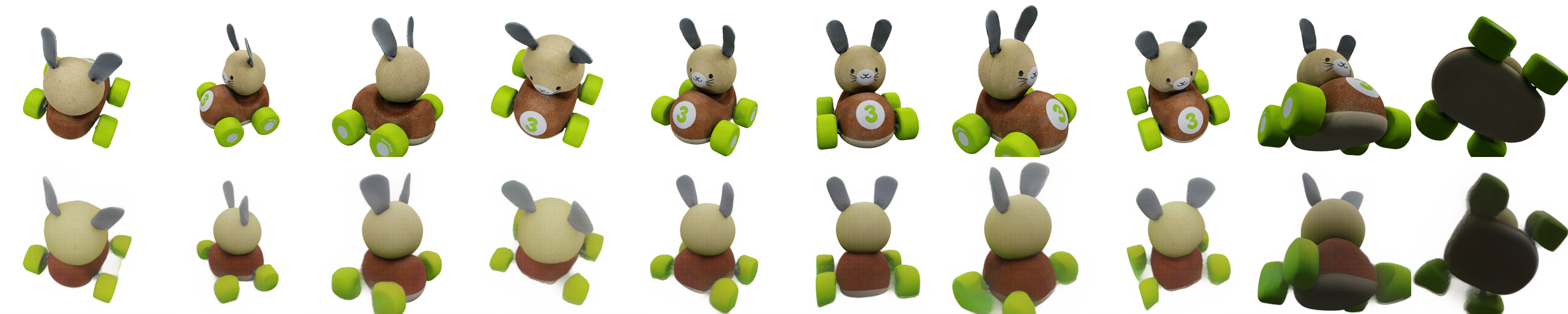} \\
\includegraphics[width=0.15\linewidth]{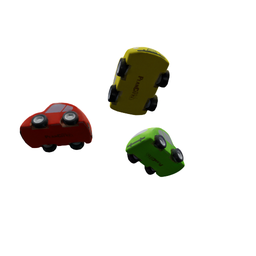} &
\includegraphics[width=0.81\linewidth]{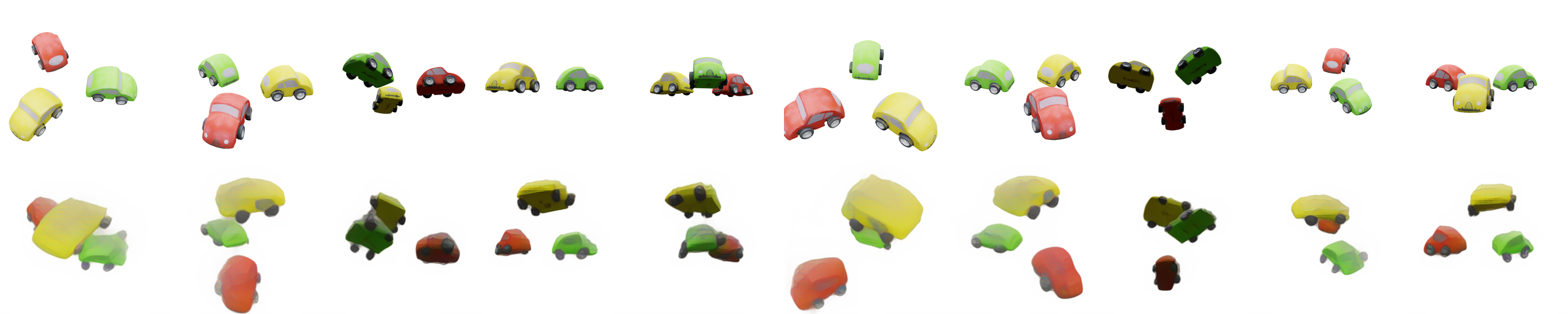} \\
\includegraphics[width=0.15\linewidth]{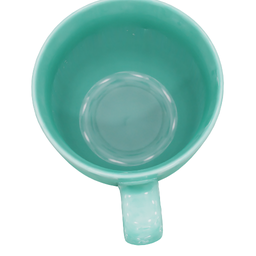} &
\includegraphics[width=0.81\linewidth]{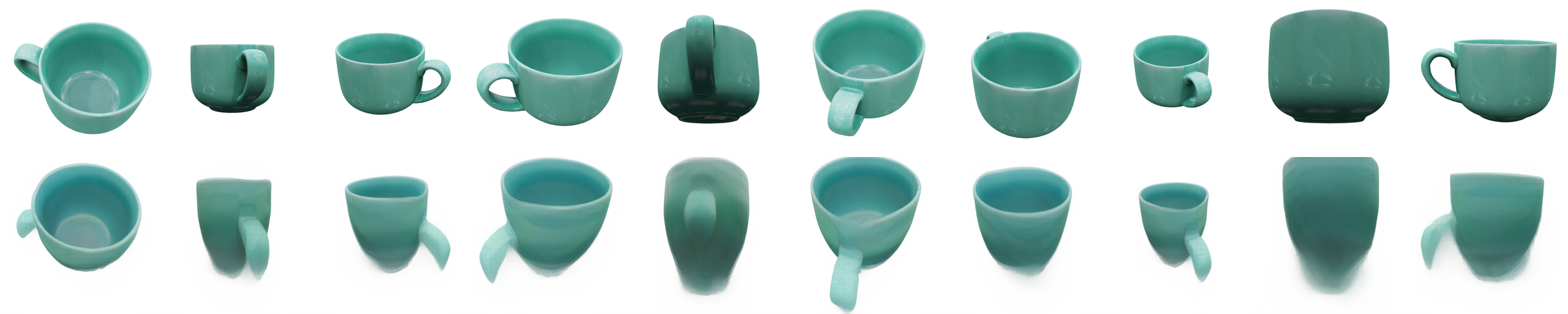} \\
\includegraphics[width=0.15\linewidth]{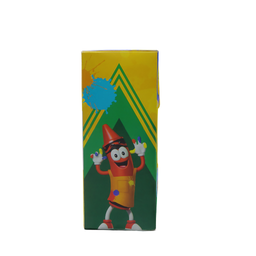} &
\includegraphics[width=0.81\linewidth]{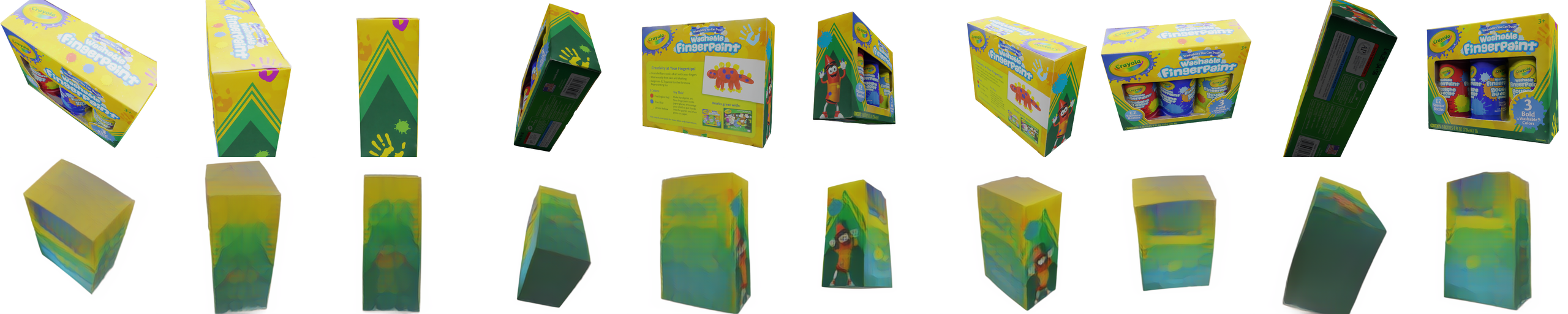} \\
source view & first rows: GT ~~~~~~~~ second rows: model prediction \\
\end{tabular}
\caption{When generating novel views from only one source image is too hard, RnG performs reasonably bad.}
\label{fig:supp_one_input_view_bad}
\end{figure*}

\begin{figure*}
\setlength{\tabcolsep}{1pt}
\small
\begin{tabular}{c|c}
\includegraphics[width=0.15\linewidth]{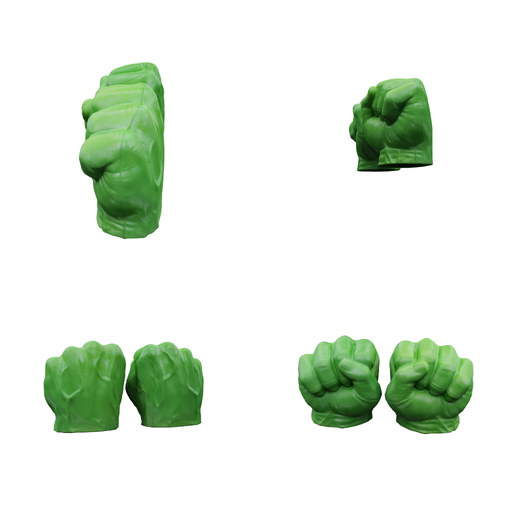} &
\includegraphics[width=0.81\linewidth]{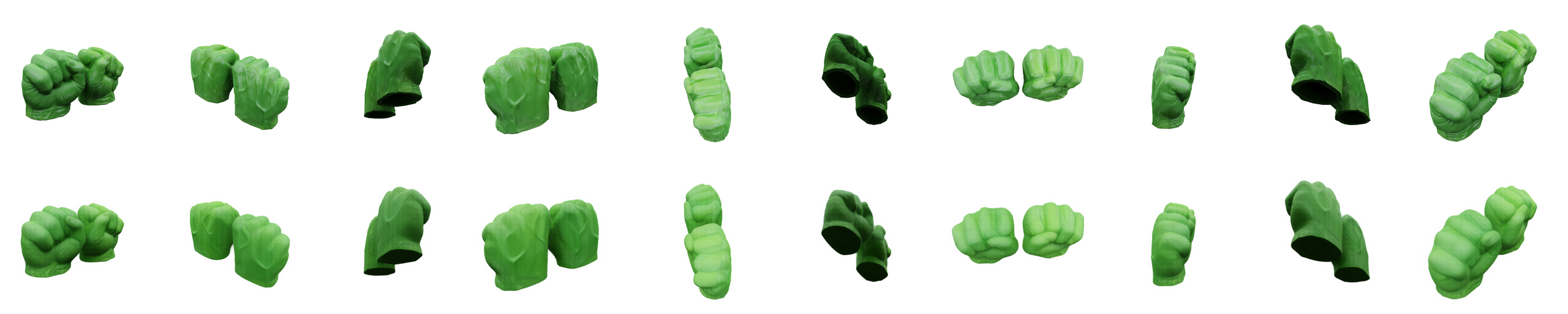} \\
\includegraphics[width=0.15\linewidth]{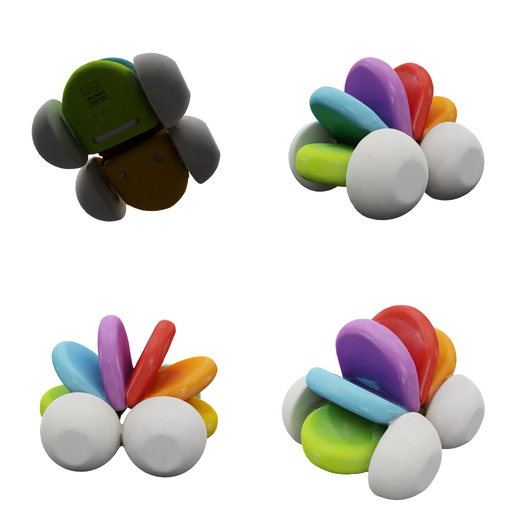} &
\includegraphics[width=0.81\linewidth]{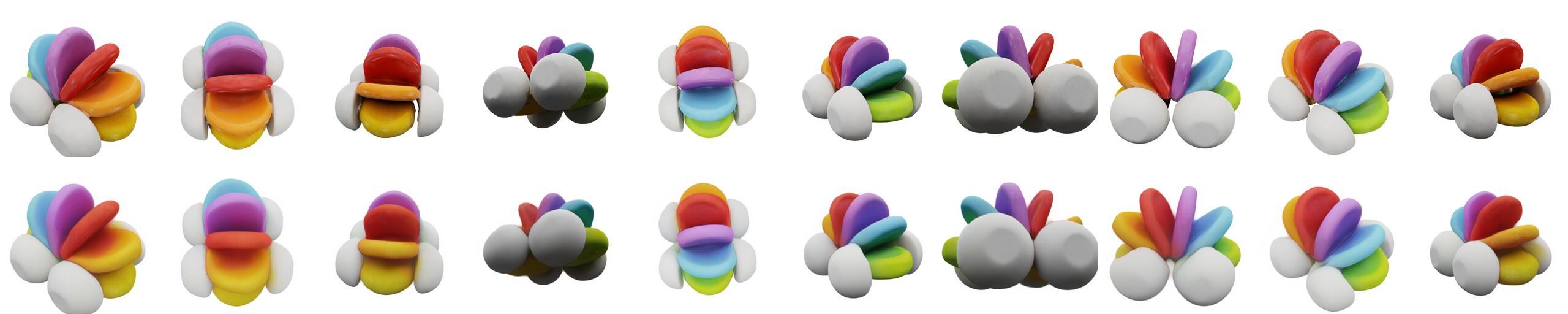} \\
\includegraphics[width=0.15\linewidth]{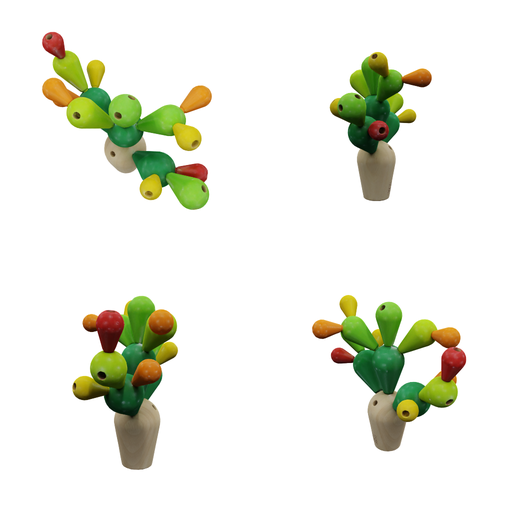} &
\includegraphics[width=0.81\linewidth]{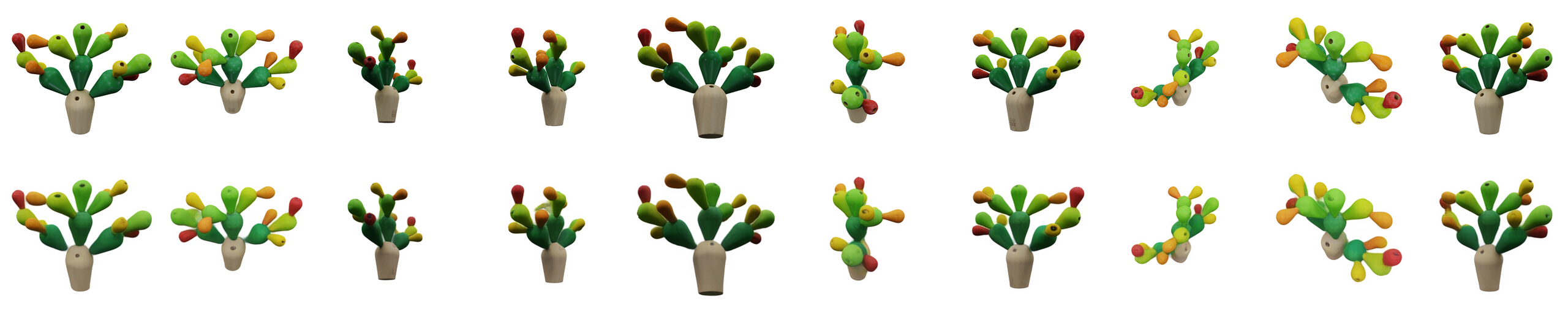} \\
\includegraphics[width=0.15\linewidth]{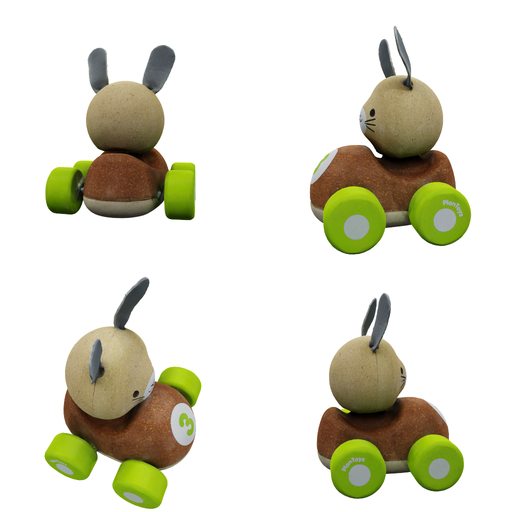} &
\includegraphics[width=0.81\linewidth]{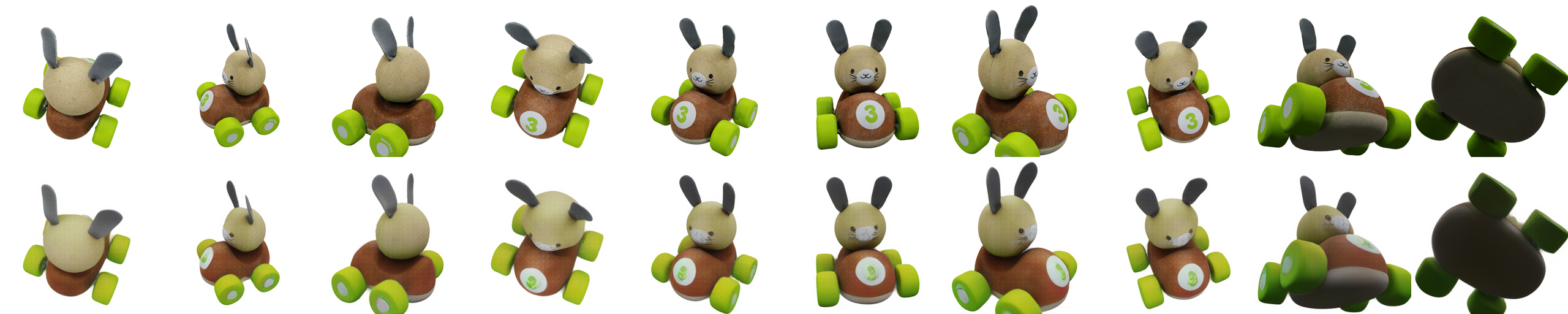} \\
\includegraphics[width=0.15\linewidth]{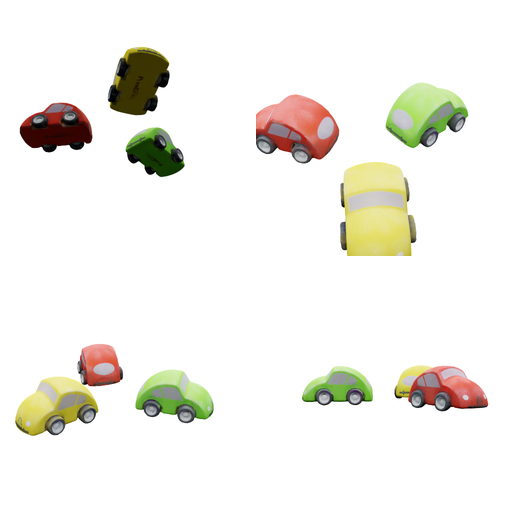} &
\includegraphics[width=0.81\linewidth]{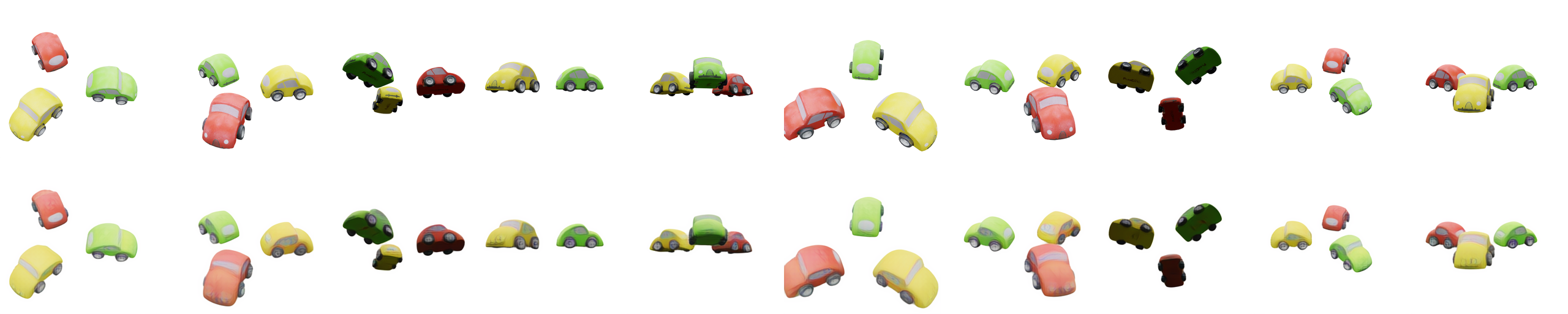} \\
\includegraphics[width=0.15\linewidth]{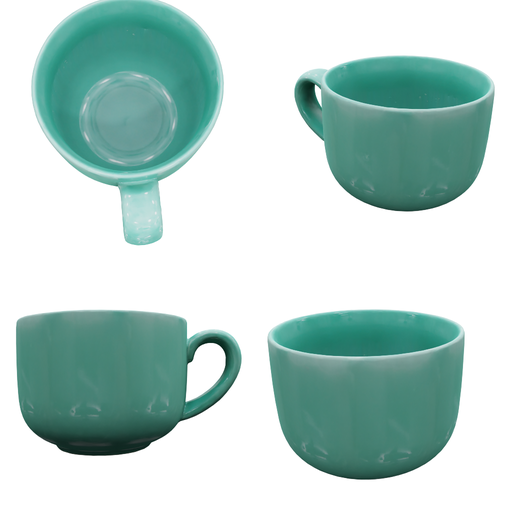} &
\includegraphics[width=0.81\linewidth]{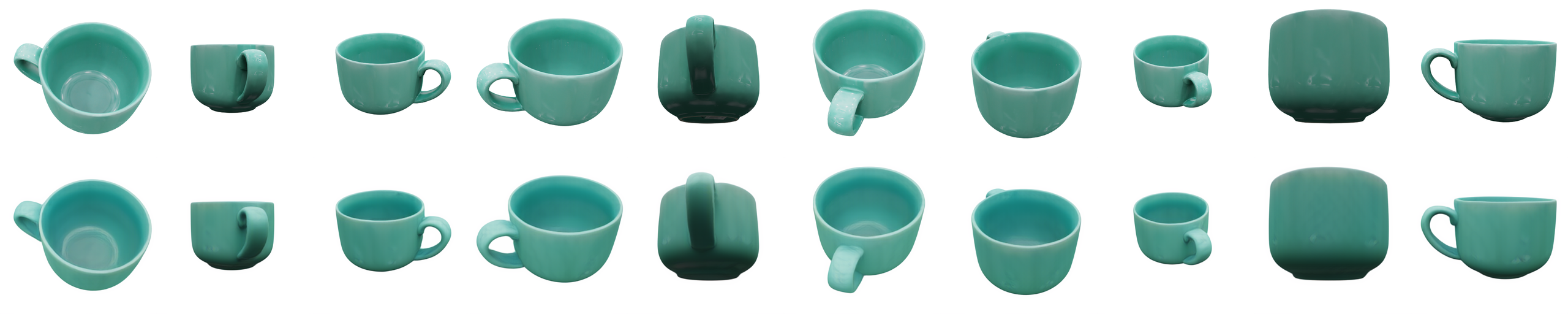} \\
\includegraphics[width=0.15\linewidth]{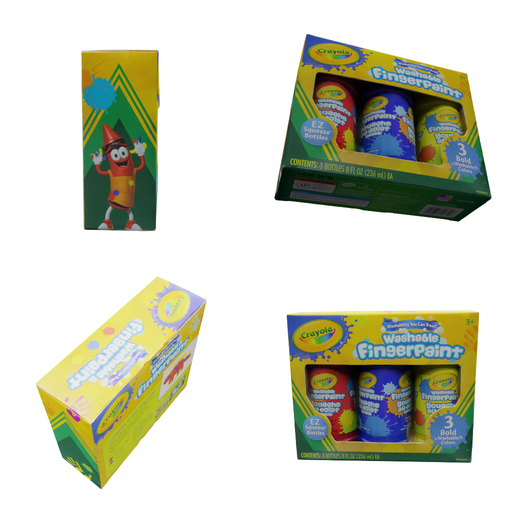} &
\includegraphics[width=0.81\linewidth]{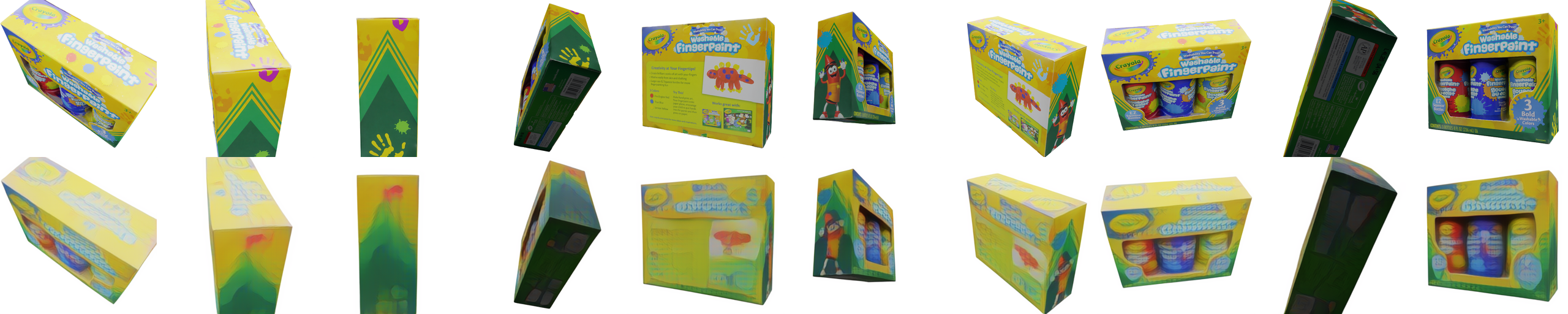} \\
source views & first rows: GT ~~~~~~~~ second rows: model prediction \\
\end{tabular}
\caption{RnG performs better when ambiguities are resolved.}
\label{fig:supp_4_input_view}
\end{figure*}

We also provide visualizations in~\autoref{fig:supp_eval_protocol}. 
As shown in the first rows,
the model fails to produce reasonable results with a single input view, but the performance improves notably with two input views.
When the number of source images increases and more observations become available, the synthesized target images exhibit richer details and higher visual fidelity. We provide more visualizations in~\autoref{fig:supp_one_input_view_good}. In these cases, the model performs reasonably well even when only a single source view is given. There are also hard cases where key structures are occluded or the structures are ambiguous. We show these samples in~\autoref{fig:supp_one_input_view_bad}, where RnG performs `reasonably bad'. When sufficient amount of observations are available, the performance of RnG improves over these cases, shown in~\autoref{fig:supp_4_input_view}.

\subsection{Generalize to real-world inputs}
\begin{figure}[h]
\centering
\vspace{-0.1in}
    \includegraphics[width=\linewidth]{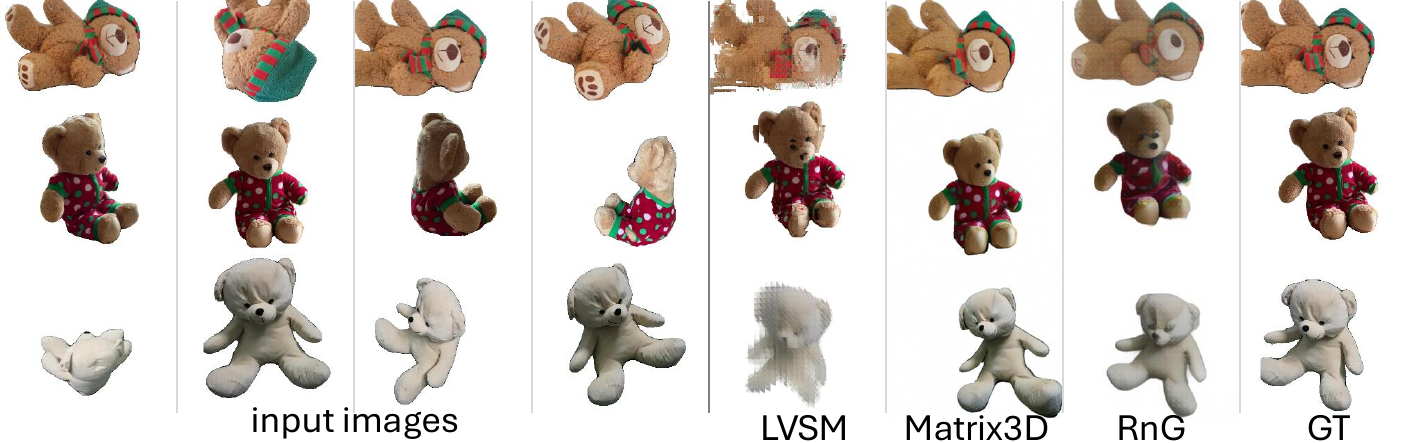}
\vspace{-0.26in}
\caption{Qualitative results on CO3D.}
\label{fig:co3d}
\end{figure}

Generalization results on CO3D are shown in~\autoref{fig:co3d}. RnG shows better zero-shot performance than LVSM and is comparable to Matrix3D trained with CO3D data. Due to the world-origin ambiguity discussed in the Limitations (\autoref{sec:limitations}), predictions may be misaligned with the ground truth, which can be alleviated by improved training data preparation. 

\subsection{Generalize to multi-objects scenerarios}
\begin{figure}[h]
\includegraphics[width=\linewidth]{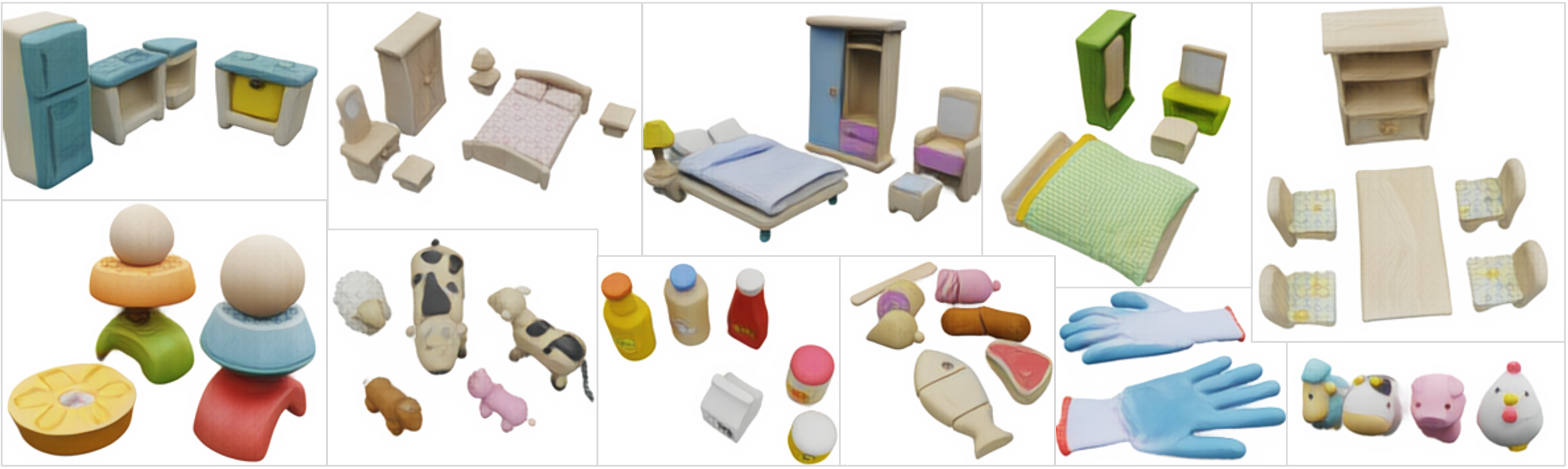}
\caption{RnG's novel view synthesis results on multi-objects.}
\label{fig:multi_obj}
\end{figure}

RnG naturally generalizes to multi-object scenes without any modification.
As shown in~\autoref{fig:multi_obj}, several examples from the GSO dataset support this claim.
We find it valuable for future explorations to systematically evaluate on more complex scenarios.

\clearpage

\end{document}